# Automatic discovery of interpretable planning strategies

Julian Skirzyński[1] · Frederic Becker[1] · Falk Lieder[1]

## Abstract

When making decisions, people often overlook critical information or are overly swayed by irrelevant information. A common approach to mitigate these biases is to provide decision-makers, especially professionals such as medical doctors, with decision aids, such as decision trees and flowcharts. Designing effective decision aids is a difficult problem. We propose that recently developed reinforcement learning methods for discovering clever heuristics for good decision-making can be partially leveraged to assist human experts in this design process. One of the biggest remaining obstacles to leveraging the aforementioned methods for improving human decision-making is that the policies they learn are opaque to people. To solve this problem, we introduce AI-Interpret: a general method for transforming idiosyncratic policies into simple and interpretable descriptions. Our algorithm combines recent advances in imitation learning and program induction with a new clustering method for identifying a large subset of demonstrations that can be accurately described by a simple, high-performing decision rule. We evaluate our new AI-Interpret algorithm and employ it to translate information-acquisition policies discovered through metalevel reinforcement learning. The results of three large behavioral experiments showed that providing the decision rules generated by AI-Interpret as flowcharts significantly improved people's planning strategies and decisions across three different classes of sequential decision problems. Moreover, our fourth experiment revealed that this approach is significantly more effective at improving human decision-making than training people by giving them performance feedback. Finally, a series of ablation studies confirmed that our AI-Interpret algorithm was critical to the discovery of interpretable decision rules and that it is ready to be applied to other reinforcement learning problems. We conclude that the methods and findings presented in this article are an important step towards leveraging automatic strategy discovery to improve human decision-making. The code for our algorithm and the experiments is available at https://github.com/RationalityEnhancement/InterpretableStrategyDiscovery.

# 1 Introduction

Human decision-making is plagued by many systematic errors that are known as cognitive biases (Gilovich et al., 2002; Tversky & Kahneman, 1974). To mitigate these biases, professionals, such as medical doctors, can be given decision aids such as decision trees and flowcharts, that guide them through a decision process that considers the most important information (Hafenbrädl et al., 2016; Laskey & Martignon, 2014; Martignon, 2003). To be practical in real-life, the strategies suggested by decision aids have to be simple (Gigerenzer, 2008; Gigerenzer & Todd, 1999) and mindful of the decision-maker's valuable time and the constraints on what people can and cannot do to arrive at a decision (Lieder & Griffiths, 2020). Previous research has identified a small set of simple heuristics that satisfy these criteria and work well for specific decisions (Gigerenzer, 2008; Gigerenzer & Todd, 1999; Lieder & Griffiths, 2020). In principle, this approach could be applied to help people in a wide range of different situations but discovering clever strategies is very difficult. Our recent work suggests that this problem can be solved by leveraging machine learning to discover near-optimal strategies for human decision-making automatically (Lieder et al., 2018, 2019, 2020; Lieder & Griffiths, 2020) (see Sect. 2). Equipped with an automatically discovered decision strategy, we may tackle many real-world problems for which there are no existing heuristics, but which are nevertheless crucial for practical applications. This includes designing decision aids for choosing between multiple alternatives (e.g., investment decisions; see Fig. 2) as well as strategies for planning a sequence of actions (e.g., a road trip, a project, or the treatment of a medical illness). For instance, a strategy for planning a road trip would help people to decide which potential destinations and pit stops to collect more information about depending on what they already know and to recognize when they have done enough planning.

One of the biggest remaining challenges is to formulate the discovered strategies in such a way that people can readily understand and apply them. This is especially problematic when strategies are discovered in the form of complex stochastic black-box policies. Here, we address this problem by developing an algorithm that approximates complex decision-making policies discovered through reinforcement learning by simple human-interpretable rules. To achieve this, we first cluster a large number of demonstrations of the complex policy and then find the largest subset of those clusters that can be accurately described by simple, high-performing decision rules. We induce those rules using Bayesian imitation learning (Silver et al., 2019). As illustrated in Fig. 1, the resulting algorithm may be incorporated into the reinforcement learning framework for automatic strategy discovery, enabling us to automatically discover flowcharts that people can follow to arrive at better decisions. We evaluated the interpretability of the decision rules our method discovered for three different three-step decision problems in large-scale online experiments. In each case people understood the rules discovered by our method and were able to successfully apply them. Importantly, another large-scale experiment revealed that the flowcharts generated by our machine learning powered approach are more effective at improving human decision-making than the status quo (i.e., training people by giving them feedback on their performance and telling them what the right decision would have been when they make a mistake).

We start with describing the background of our approach in Sect. 2 and present our problem statement in Sect. 3. Section 4 focuses on related work. In Sect. 5, we introduce a new approach to interpretable RL—AI-Interpret—along with a pipeline for generating decision aids through automatic strategy discovery. In Sect. 6, using behavioral experiments, we demonstrate that decision aids designed with the help of automatic strategy discovery and AI-Interpret can significantly improve human decision-making. The results in Sect. 7 show that AI-Interpret was critical to this success. We close by discussing potential real-world applications of our approach and directions for future work.

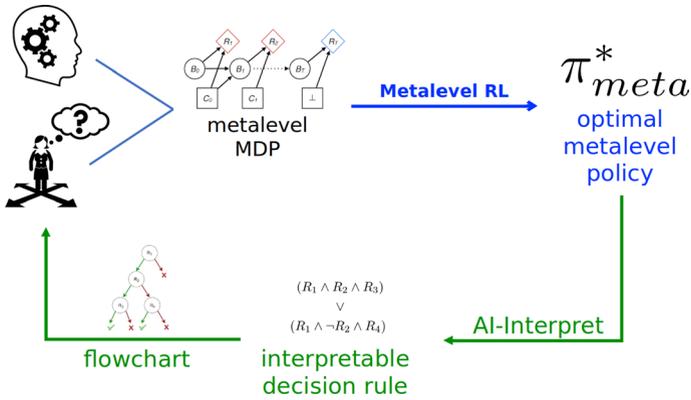

**Fig. 1** Our framework for improving human decision-making through automatic strategy discovery (Lieder et al., 2019, 2020). As illustrated in the upper row, the approach starts with modeling the decision problems people face in everyday life and how they can make those decisions in the framework of metalevel MDPs (Callaway et al., 2018b, 2019; Griffiths et al., 2019; Lieder et al., 2017, 2018, 2019, 2020; Lieder & Griffiths, 2020). The optimal algorithm for human decision-making can be discovered by computing the optimal metalevel policy through metalevel reinforcement learning (Callaway et al., 2018a; Lieder et al., 2017). The contribution of this paper is to develop an algorithm called AI-Interpret that translates the resulting metalevel policies into flowcharts that people can follow to make better decisions

## 2 Background

In this section, we define the formal machinery that the methods presented in this article are based on. We start with the basic framework for describing decision problems (i.e., Markov Decision Processes). We then proceed to present a formalism for the problem of deciding how to decide (i.e., metalevel Markov Decision Processes). The possible solutions to the problem of deciding how to decide are known as *decision strategies*. Our approach to improving human decision making through automatic strategy discovery rests on a mathematical definition of what constitutes an optimal strategy for human decision-making known as *resource-rationality*. We introduce this theory in the third part of this Background section. Then, we briefly review existing methods for solving metalevel MDPs and move to the topic of imitation learning to describe the family of methods that our algorithm for interpretable RL belongs to. Afterwards, we define disjunctive normal form formulas which constitute the formal output of our algorithm. We finish with describing a baseline imitation learning method we built on.

### 2.1 Modeling sequential decision problems

In general, AI-Interpret considers reinforcement learning policies defined on finite Markov Decision Processes. A Markov Decision Process (MDP) is a formal framework for modeling sequential decision problems. In a sequential decision problem an agent (repeatedly) interacts with its environment. In each of a potential long series of interactions the agent observes the state of the environment and then selects an action that changes the state of the environment and generates a reward.

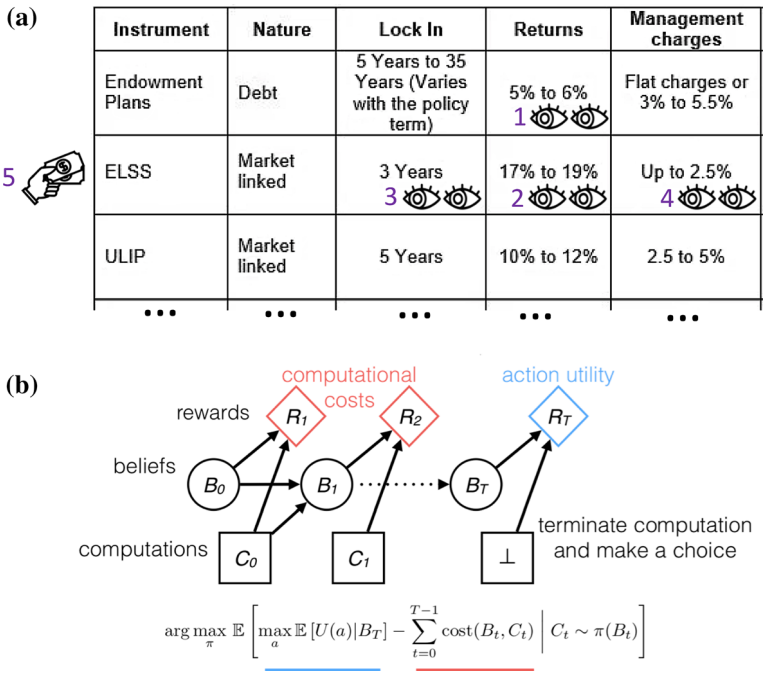

**Fig. 2** Formalizing the optimal algorithms for human decision-making in the real-world as the solution to a metalevel MDP. a) Illustration of a real-life decision-problem and an efficient heuristic decision strategy for making such decisions. In this example, the goal is to choose between multiple investment options based on several attributes. Critically, such decision problems and the optimal strategies for making such decisions can be formalized in the computational framework of metalevel MDPs illustrated in Panel b. The decision-maker's goal can be modelled as maximizing the expected value of the chosen option minus the time cost of making the decision (Lieder & Griffiths, 2020). The expected subjective value of an alternative $a$ given the acquired information $B_T$ at the time of the decision ($T$) can be modelled as a weighted sum of its scores on several attributes (e.g., $\mathbb{E}[U(a)|B_T] = 0.4 \cdot \text{Returns}(a) - 0.2 \cdot \text{ManagementCharges}(a) - 0.3 \cdot \text{LockIn}(a) + 0.05 \cdot \text{IsMarketLinked}$ where the weights reflect the investor's preferences). To estimate the alternatives' subjective values, the decision-maker has to perform computations $C$ by acquiring information (e.g., the management charges for a given investment) and updating its beliefs ($B$) accordingly. Each computation has a cost ($\text{cost}(B_T, C_t)$). The optimal decision strategy maximizes the expected subjective value of the final decision minus the cumulative cost of the decision operations that had to be performed to reach that decision. The sequence of computations suggested in Panel a) is a demonstration of a decision strategy that exploits the decision-maker's preferences and prior knowledge about the distribution of attribute values to minimize the amount of effort required to identify the best option with high probability. This is achieved by using the most important attribute (Returns) to decide which alternatives to eliminate, which alternative to examine more closely, and when to stop looking for better alternatives. b) Illustration of a metalevel MDP (see Definition 2). A metalevel MDP is a Markov Decision Process where actions are computations ($C$) and states encode the agent's beliefs ($B$). The rewards for computations ($R_1, R_2, \ldots$) measure the cost of computation and the reward for terminating deliberation ($R_T$) is the expected return for executing the plan that is best given the current belief state ($B_T$). Discovering the optimal strategies corresponds to computing the optimal metalevel policy, which achieves an optimal trade-off between decision quality and computational cost

**Definition 1** (*Markov decision process*) A Markov decision process (MDP) is a finite process that satisfies the Markov property (each state in the process is independent of the history). It is represented by a tuple $(\mathcal{S}, \mathcal{A}, \mathcal{T}, \mathcal{R}, \gamma)$ where $\mathcal{S}$ is a set of states; $\mathcal{A}$ is a set of actions; $\mathcal{T}(s, a, s') = \mathbb{P}(s_{t+1} = s' \mid s_t = s, a_t = a)$ for $s \neq s' \in \mathcal{S}, a \in \mathcal{A}$ is a state transition function; $\gamma \in (0, 1)$ is a discount factor; $\mathcal{R} : \mathcal{S} \to \mathbb{R}$ is a reward function.

Note that $\mathcal{R}$ could be also represented as a function dependent on state-action pairs $\mathcal{R} : \mathcal{S} \times \mathcal{A} \to \mathbb{R}$. Policy $\pi : \mathcal{S} \to \mathcal{A}$ denotes a deterministic function that controls agent's behavior in an MDP and a nondeterministic $\pi : \mathcal{S} \to Prob(\mathcal{A})$ defines a probability distribution over the actions. The cumulative return of a policy is a sum of its discounted rewards, i.e. $G_t^\pi = \sum_{i=t}^{\infty} \gamma^t r_t$ for $\gamma \in [0, 1]$.

### 2.2 Modeling the problem of deciding how to decide as a metalevel MDP

Computing the optimal policy for a given MDP corresponds to planning a sequence of actions. Optimal planning quickly becomes intractable as the number of states and the number of time steps increase. Therefore, people and agents with performance-limited hardware more generally, have to resort to approximate decision strategies. Decision strategies differ in which computations they perform in which order depending on the problem and the outcomes of previous computations. Each computation may generate new information that allows the agent to update its beliefs about how good or bad alternative courses of action might be. Here, we focus on the case where the transition matrix $T$ is known and the computations reveal the rewards $r(s, a)$ of taking action $a$ in state $s$. In this context, the agent's belief $b_t$ at time $t$ can be represented by a probability distribution $P(\mathbf{R})$ on the entries $\mathbf{R}_{s,a} = r(s, a)$ of the reward matrix $\mathbf{R}$. Computation is costly because the agent's time and computational resources are limited. The problem of deciding which computations should be performed in which order can be formalized as a metalevel Markov Decision Process (Hay et al., 2012; Griffiths et al., 2019). A metalevel decision process is a Markov Decision Process (see Definition 1) where the states are beliefs and the actions are computations.

**Definition 2** (*Metalevel Markov decision process*) A metalevel MDP (Hay et al., 2012; Griffiths et al., 2019) is a finite process represented by a 4-tuple $(\mathcal{B}, \mathcal{C}, \mathcal{T}_{meta}, \mathcal{R}_{meta})$ where $\mathcal{B}$ is a set of beliefs; $\mathcal{C}$ is a set of computational primitives; $\mathcal{T}_{meta}(b, c, b') = \mathbb{P}(b_{t+1} = b' \mid b_t = b, c_t = c)$ is a belief transition function; $\mathcal{R}_{meta} : \mathcal{C} \cup \{\perp\} \to \mathbb{R}$ is a reward function which captures the cost of computations in $\mathcal{C}$ and the utility of the optimal course of actions after terminating the computations with $\perp$.

Analogically to the previous case, $\pi_{meta} : \mathcal{B} \to \mathcal{C}$ is a deterministic metalevel policy that controls how the agent is making computations, and $\pi_{meta} : \mathcal{B} \to Prob(\mathcal{C})$ defines a probability distribution over the computations. One can think of metalevel policies as mathematical descriptions of decision strategies that a person could use to choose between multiple alternatives or plan a sequence of actions.

### 2.3 Resource-rationality

Our approach to discovering decision strategies that people can use to make better choices is rooted in the theory of resource-rationality (Lieder & Griffiths, 2020). Its basic idea is that people should use cognitive strategies that make the best possible use of their finite time and bounded cognitive resources. The optimal decision strategy critically depends on the structure of the environment $E$ that the decision maker is interacting with and the computational resources afforded by their brain $B$. These two factors jointly determine the quality of the decision that would result from using a potential decision strategy $h$ and how costly it would be to apply this strategy. The resource-rationality of a strategy is the net benefit of using a decision strategy minus its cost.

**Definition 3** (*Resource-rationality*) The resource-rationality (RR) of using the decision strategy $h$ in the environment $E$ is

$$\text{RR}(h, E, B) = \mathbb{E}_{P(result|s_0,h,E,B)}[u(result)] - \mathbb{E}_{P(t_h,\rho|s_0,h,E,B)}[cost(t_h, \rho)], \qquad (1)$$

where $s_0 = (o, b_0)$ comprises what the agent's observations about the environment ($o$) and its internal state $b_0$, $u$(result) is the agent's utility of the outcomes of the decisions that the strategy $h$ might make, and $cost(t_h, \rho)$ denotes the total opportunity cost of investing the cognitive resources $\rho$ used or blocked by strategy $h$ until it terminates deliberation at time $t_h$. Expectations are taken with respect to the posterior probability distribution of possible results given the environment $E$ and the agent's observations $o$ about its current state.

Note that the execution time and the possible results of the strategy depend on the situation in which it is applied.

**Definition 4** (*Resource-rational strategy*) A strategy $h^\star$ is said to be resource-rational for the environment $E$ under the limited computational resources that are available to the agent if

$$h^\star = \arg\max_{h \in H_B} \mathbb{E}_E[RR(h, E, B)], \qquad (2)$$

that is when $h^*$ maximizes the value of resource-rationality among all the strategies $H_B$ that the agent can execute.

Discovering resource-rational strategies can be expressed as a problem of finding the optimal policy for a metalevel MDP where the states represent the agent's beliefs, the actions represent the agent's computations, and the rewards are inherited from the costs of computations and the value of terminating under the current belief state. It is possible to solve for this strategy using dynamic programming (Callaway et al., 2019) and reinforcement learning (Callaway et al., 2018a; Lieder et al., 2017).

### 2.4 Solving metalevel MDPs

Computing effective decision strategies and modeling the problem of which decision procedure people should use entails using exact or approximate MDP-solvers on metalevel MDPs. Exact methods for solving MDPs, such as dynamic programming, can be applied to small metalevel MDPs. But as the size of the environment increases, dynamic programming quickly becomes intractable. Therefore, the primary methods for solving metalevel MDPs are reinforcement learning (RL) algorithms that approximate the optimal metalevel policy $\pi^*_{meta}$ through trial and error (Callaway et al., 2018a; Lieder et al., 2017; Kemtur et al., 2020). The primary shortcoming of most methods for solving (metalevel) MDPs is that the resulting policies are very difficult to interpret. Our primary contribution is to develop a method that makes them interpretable. To achieve this, our method performs *imitation learning*.

## 2.5 Imitation learning

Imitation learning (IL) is the problem of finding a policy $\hat{\pi}$ that mimicks transitions provided in a dataset of trajectories $\mathcal{D} = \{(s_i, a_i)\}_{i=1}^{M}$ where $s_i \in \mathcal{S}, a_i \in \mathcal{A}$ (Osa et al., 2018). Unlike canonical applications of IL that focus on imitating behavioral policies, our application focuses on metalevel policies for selecting computations.

## 2.6 Boolean logics

The formal output of the algorithm we introduce is a logical formula in disjunctive normal form (DNF).

**Definition 5** (*Disjunctive normal form*) Let $f_{i,j}, h : \mathcal{X} \to \{0, 1\}, i, j \in \mathbb{N}$ be binary-valued functions (predicates) on domain $\mathcal{X}$. We say that $h$ is in disjunctive normal form if the following property is satisfied:

$$h(x) = (f_{1,1}(x) \wedge \cdots \wedge f_{1,n_1}((x)) \vee \cdots \vee (f_{m,1}((x)) \wedge \cdots \wedge f_{m,n_m}((x)) \tag{3}$$

and $\forall i, j_1 \neq j_2, f_{i,j_1} \neq f_{i,j_2}$.

Equality 3 defines that a predicate is in DNF if it is a disjunction of conjunctions and every predicate appears only once in each conjunction.

## 2.7 Logical program policies

Logical Program Policies (LPP) is a Bayesian imitation learning method that given a set of demonstrations $\mathcal{D} = \{(s_i, a_i)\}_{i=1}^{M}$, outputs a posterior distribution over logical formulas in disjunctive normal form (DNF; see Definition 5) that best describe the generated data. For that purpose, the authors restrict the considered set of solutions to formulas $\{h_1(s, a), \ldots, h_n(s, a)\}$ defined in a domain-specific language (DSL)—a set of predicates $f_i(s, a) : \mathcal{S} \times \mathcal{A} \to \{0, 1\}$, called (simple) *programs*. Programs are understood as feature detectors which assign truth values to state-action pairs, and formulas over programs are the titular *logical programs*. To find the best $K$ logical programs $h_i^{\text{MAP}}$, the authors employ maximum a posteriori estimation (MAP). Each program $h_i^{\text{MAP}}$ induces a stochastic policy

$$\pi_i^{\text{MAP}}(s, a) = \frac{h_i^{\text{MAP}}(s, a)}{\sum_{a'} h_i^{\text{MAP}}(s, a')} \tag{4}$$

which is a uniform distribution over all actions $a$ for which $h_i^{\text{MAP}}$ is true in state $s$. The program-level policy $\hat{\pi}$ integrates out the uncertainty about the possible programs and selects the action

$$\hat{\pi}(s) = \arg\max_{a \in \mathcal{A}} \sum_i q(h_i^{\text{MAP}}) \cdot \pi_i^{\text{MAP}}(s, a). \tag{5}$$

Importantly for our application, the formulas which constitute the program-level policy come in disjunctive-normal form. To obtain them, the set of demonstrated state-action pairs $(s_i, a_i) \in \mathcal{D}$ is used to automatically create binary feature vectors $v_i^+ = \langle f_1(s_i, a_i), \ldots, f_m(s_i, a_i)\rangle$ and binary feature vectors $v_i^- = \langle f_1(s_i, a_j), \ldots, f_m(s_i, a_j)\rangle$, $j \neq i$. The $v_i^+$ vectors serve as as positive examples and the $v_i^-$ vectors, which describe non-demonstrated actions in observed states, as negative examples. After applying an off-the-shelf decision tree induction method (e.g. Pedregosa et al., 2011; Valiant, 1985) on all $v_i^+$s and $v_i^-$s, the formulas are extracted from the tree by treating each path leading to a positive decision as a conjunction of predicates. Considering all positive decision paths as an alternative results in a DNF formula. Further in the text, the notation *LPP*($\mathcal{D}$) will stand for the formula generated by Logical Program Policies method on the set of demonstrations $\mathcal{D}$.

## 3 Problem definition and proposed solution

The main goal of the presented research is to develop a method that takes in a model of the environment and a model of people's cognitive architecture and returns a verbal or graphical description of an effective decision strategy that people will understand and use to make better decisions. This problem statement goes beyond the standard formulation of interpretable AI by requiring that when people are given an "interpretable" description of a decision strategy they can execute that strategy themselves. Because of that, we employ a novel algorithm evaluation approach in our work, which has not been utilized in standard interpretablility research. Concretely, we measure how much the generated descriptions boost the performance of human decision makers in simulated sequential decision problems.

Our approach to discovering human-interpretable decision strategies comprises three steps: (1) formalizing the problem of strategy discovery as a metalevel MDP (see Fig. 2, Sect. 2.2), (2) developing reinforcement learning methods for computing optimal metalevel policies (see Sect. 2.4), and (3) describing the learned policies by simple and human-interpretable flowcharts. In previous work, we proposed solutions to the first two sub-problems. First, we conceptualized the strategy discovery as a problem in the realm of metalevel reinforcement learning (Lieder et al., 2017, 2018; Callaway et al., 2018a; Griffiths et al., 2019) as described in Sect. 2.2. Second, we introduced methods to find optimal policies for metalevel MDPs. Our Bayesian Metalevel Policy Search (BMPS) algorithm (Callaway et al., 2018a, 2019; Lieder & Griffiths, 2020) relies on the notion of the value of computation *VOC*. *VOC* is the expected improvement in decision quality that can be achieved by performing computation $c$ in belief state $b$ (and continuing optimally) minus the cost of the optimal sequence of computations. BMPS approximates *VOC*($c, b$) by a linear combination of features that are estimated from the original metalevel MDP through agent's interaction with the environment. Knowing that $\pi^* = \arg\max_{c \in C} VOC(c, b)$, by approximating *VOC*, BMPS finds a near-optimal metalevel policy. Apart from BMPS, we also developed a dynamic programming method for finding exact solutions for fairly basic metalevel MDPs (Callaway et al., 2020). In this work, we turn to the third sub-problem: constructing simple human-interpretable descriptions of the automatically discovered decision strategies. Our goal is to develop a systematic method for transforming complex, black-box policies into simple human-interpretable flowcharts that allow people to approximately follow the near-optimal decision strategy expressed by the black-box policy. As a proof of concept, we study this problem in the domain of planning.

To handle an arbitrary type of an RL policy, thereby generalizing beyond metalevel RL, and simplify the creation of flowcharts, we propose to use imitation learning with DNF formulas. Imitation learning can be applied to arbitrary policies. Moreover, the demonstrations it needs may be gathered by executing the policy either in a simulator or in the real world. Obtaining a DNF formula, in turn, allows one to describe the policy by a decision tree. This representation is a strong candidate for interpretable descriptions of procedures (Gigerenzer, 2008; Hafenbrädl et al., 2016) and may be easily transformed to a flowchart. Formally, we start with a set of demonstrations $\mathcal{D} = \{(s_i, a_i)\}_{i=1}^{M}$ generated by policy $\pi$, and denote $I(f) = \sum_{i=1}^{M} \mathbb{1}_{\pi_f(s_i) = a_i}$ as the number of demonstrations that the policy induced by formula $f$ can imitate properly. Let's denote the set of all DNF formulas defined in a domain-specific language $\mathcal{L}$ whose longest conjunction is of length $d$ as $DNF(\mathcal{L}, d)$. Let's also assume that policy $\pi_a$ is $\alpha$-similar to policy $\pi_b$ if the expected return of $\pi_a$: $\mathbb{E}(G_0^{\pi_a})$ constitutes at least $\alpha$ of the expected return of $\pi_b$, i.e. $\mathbb{E}(G_0^{\pi_a})/\mathbb{E}(G_0^{\pi_b}) \geq \alpha$. We denote $F_\alpha^{d,\mathcal{L}} = \{f \in DNF(\mathcal{L}, d) : \frac{\mathbb{E}(G_0^{\pi_f})}{\mathbb{E}(G_0^{\pi})} \geq \alpha\}$ as the set of all DNF formulas with conjunctions smaller than $d$ predicates that induce policies $\alpha$-similar to $\pi$. We would like to find a formula $f^*$ that belongs to $F_\alpha^{d,\mathcal{L}}$ for arbitrary $\alpha$ and $d$, maximizing the number of demonstrations that $\pi_{f^*}$ can imitate, that is

$$f^* = \arg\max_{f \epsilon F_\alpha^{d,L}} I(f). \tag{6}$$

To approximate the optimization process above-specified, we develop a general method for creating descriptions of RL policies: the Adaptive Imitation-Interpretation algorithm (AI-Interpret). Our algorithm captures the essence of the input policy by finding its simpler representation (small $d$) which performs almost as well (small $\alpha$). To accomplish that, AI-Interpret builds on the Logical Program Policies (LPP) method by Silver et al. (2019). Similarly to the latter, AI-Interpret accepts a set of demonstrations of the policy $\mathcal{D}$ and $\mathcal{L}$, a domain-specific language (DSL) of predicates which captures features of states that could be encountered and actions that could be taken in the environment under consideration. AI-Interpret uses the constructed predicates to separate the set of demonstrations into clusters. Doing so, enables it to consider increasingly smaller sets of demonstrations and employ LPP in a structured search for simple logical formulas in $F_\alpha^{d,\mathcal{L}}$. To improve human decision making, we use AI-Interpret in our strategy discovery pipeline (see Fig. 1). After modeling the planning problem as a metalevel MDP, using RL algorithms to compute its optimal policy, gathering this policy's demonstrations, and creating a custom DSL, the pipeline uses AI-Interpret to find a set of candidate formulas. The formulas are transformed into decision trees, and then visualized as flowcharts that people can follow to execute the strategy in real life.

## 4 Related work

*Strategy discovery* Historically, discovering strategies that people can use to make better decisions and developing training programs and decision aids that help people execute such strategies was a manual process that relied exclusively on human expertise (Hafenbrädl et al., 2016; Laskey & Martignon, 2014; Martignon, 2003). Recent work has been increasingly more concerned with discovering human decision strategies automatically

(Binz & Endres, 2019; Callaway et al., 2018a, 2018b, 2019; Kemtur et al., 2020; Lieder et al., 2017, 2018) and with using cognitive tutors (Lieder et al., 2019; Lieder et al., 2020). In our previous studies on this topic (Binz & Endres, 2019; Callaway et al., 2018a, 2018b, 2019; Kemtur et al., 2020; Lieder et al., 2017, 2018), we enabled automatic discovery of optimal decision strategies through leveraging reinforcement learning. In technical terms, we formalized decision strategies that make the best possible use of the decision-maker's precious time and finite computational resources (Lieder & Griffiths, 2020) within the framework of metalevel MDPs (see Fig. 2 and Sect. 2.2). This approach, however, led to stochastic black-box metalevel policies whose behavior can be idiosyncratic.

*Visual representations of RL Policies* Most approaches to interpretable AI in the domain of reinforcement learning try to describe the behavior of RL policies visually. Lam et al. (2020), for instance, employed search tree descriptions. The search trees had a graphical representation which contained current and expected states, win probabilities for certain actions, and connections between the two. They were learned similarly as in AlphaZero (Silver et al., 2018), through Q function optimization and self-play, and by additionally learning the dynamics model. Lam et al. (2020) studied an interactive, sparse-reward game Tug of War and showed how domain experts may utilize search trees to verify policies for that game. Annasamy and Sycara (2019) proposed the i-DQN algorithm that learned a policy alongside its interpretation in form of images with highlighted elements representative to the decision-making. They achieved that by constraining the latent space to be reconstructible through key-stores, vectors encoding the model's global attention over multiple states and actions. Their tests point out that inverting the key-stores provides insights into the features learned by the RL model. RL policies represented visually in other ways, for instance as attention or saliency maps, can be also found in Mott et al. (2019), Atrey et al. (2019), Greydanus et al. (2018), Puri et al. (2019), Iyer et al. (2018), Yau et al. (2020).

*Decision trees* Visual representations may also take a descriptive form of a decision tree, an idea we explored in this paper. For example, Liu et al. (2019) approximated a neural policy using an on-line mimic learner algorithm that returns Linear Model U-Trees (LMUT). LMUT represent Q-functions for a given MDP in a regression decision tree structure (Breiman et al., 1984) and are learned using the stochastic gradient descent algorithm. By extracting rules from the learned LMUT, it is possible to comprehend action decisions for a given state considering conditions that are imposed on its feature-representation. Silva et al. (2020) went even further and introduced RL function approximation technique performed via differentiable decision trees. Their method outperformed neural networks after both were learned with gradient descent, and returned deterministic descriptions that were more interpretable than networks or rule lists. Similar approaches to mimic neural networks with tree structures were presented in Bastani et al. (2018), Alur et al. (2017), Che et al. (2016), Coppens et al. (2019), Jhunjhunwala (2019), Krishnan et al. (1999).

*Imitation learning* Alternative approaches attempt to describe policies in formalisms other than decision trees. In any case, however, the interpretability methods often employ imitation learning. Notably, the Programmatically Interpretable Reinforcement Learning (PIRL) method, proposed by Verma et al. (2018) combines deep RL with an efficient program policy search. PIRL defines a domain-specific language of atomic statements with an imposed syntax and allows describing neural network-based policies with programs. These programs mimic the policies and may be found using imitation learning methods. Experiments in TORCS car-racing environment (Wymann et al., 2015) showed that this approach can learn well-performing if-else programs (Verma et al., 2018). Araki et al. (2019) proposed to represent policies as finite-state automata (FSA) and introduced a method to derive interpretable transition matrices out of FSA. In their framework, Araki et al. (2019)

used expert trajectories to (i) learn an MDP modeling actions in the environment and (ii) learn transitions in an FSA of logical expressions, to then perform value iteration over both. The authors showed how their method succeeds in multiple navigation and manipulation tasks by beating standard value iteration and a convolutional network. Similar approaches that aim to learn descriptions via imitation learning were introduced in Bhupatiraju et al. (2018), Penkov and Ramamoorthy (2019), Verma (2019), Verma et al. (2019).

## 5 Algorithm for interpretability

In this section, we introduce Adaptive Imitation-Interpretation (AI-Interpret), an algorithm that transforms the policy learned by a reinforcement learning agent into a human-interpretable form (see Sect. 3). We begin with explaining what contribution AI-Interpret makes and then provide a birds-eye view of how its components—LPP and clustering—work together to produce human-interpretable descriptions. Afterwards, we detail our approach and present a heuristic method for choosing the number of demonstration-clusters used by AI-Interpret. In the last part of this section, we analyze the whole pipeline that uses the introduced algorithm to automatically discover interpretable planning strategies.

### 5.1 Technical contribution

The technical novelty of our algorithm lies in (a) giving the user control over the tradeoff between the complexity of the output policy's description and the performance of the strategy it conveys; and (b) handling cases when the created DSL is insufficient to imitate the whole set of demonstrations, saving time-consuming fine-tuning of the DSL. We enable these components by approximating the optimization from Eq. 6. Note that neither of them is available with the baseline imitation method we employed.

Firstly, in the original formulation of LPP, the user cannot control the final form of the output other than specifying its DSL, which needs to be optimized to work well with LPP. Similarly, it is unclear how well the policy induced by LPP performs in the environment in question, and how it compares to the policy that is being interpreted. In consequence, LPP generates solutions of limited interpretability.

Secondly, LPP's performance is highly sensitive to how well the set of domain predicates is paired with the set of demonstrations. Sometimes, this sensitivity prevents the algorithm from finding any solution, even though $DNF(\mathcal{L}, d)$ is non-empty for some $d \in \mathbb{N}$. Formally, the set of demonstrations $\mathcal{D}$ is equally divided into two disjoint sets $\mathcal{D}_1, \mathcal{D}_2$ where one is used for learning programs $h_1, h_2, \ldots, h_n$, and the other gives an unbiased estimate of the likelihood $\mathbb{P}(\mathcal{D} \mid h_i) \propto \mathbb{P}(\mathcal{D}_2 \mid h_i)$. If the predicates are too specific or too general, many, if not all the considered programs, could generate a likelihood of 0 for $\mathcal{D}_2$, given that they were chosen to account well for the data in $\mathcal{D}_1$. It can also happen that no formula can be found for $\mathcal{D}_1$ itself because the dataset $\mathcal{D}$ contains very rare examples of the policy's behavior (rarely encountered state-action pairs) that the predicates cannot explain. For example, consider a situation where the demonstrations describe a market-trading policy. The birttleness of the market might result in capturing an idiosyncratic behavior of the policy, one which is very local and rather rare. This could break the results for LPP. It could either be because the set of predicates cannot capture the idiosyncratic behavior or that this behavior is used for estimating the likelihood of the programs. Despite the modeler's prior knowledge, considerable optimization might be needed to obtain a set of acceptable predicates for which neither of those issues arise.

### 5.2 Overview of the algorithm

To enable innovations mentioned in the previous subsection, we introduce an adaptive manipulation of the dataset $\mathcal{D}$. Algorithm 1 revolves around LPP, but outputs an approximate solution even in situations in which LPP would not be able to find one.

---

**Algorithm 1:** Adaptive Imitation-Interpretation

**Input:** Expert dataset $\mathcal{D} = \{(s_i, a_i)\}_{i=1}^{M}$ generated by policy $\pi$;
Domain-Specific Language of predicates $\mathcal{L} = \{f_1(s,a), ..., f_n(s,a)\}$;
Aspiration value $\alpha$;
Number of rollouts $L$;
Tolerance $\delta$;
Mean reward $m$ of $\pi$;
Max size of a conjunction in the output formula $d$;
Number of clusters $N$;

**Output:** Interpretable DNF formula $f$;

1: Map state-action pairs in $\mathcal{D}$ to vectors of truth values:
   $(s_i, a_i) \mapsto \langle f_1(s_i, a_i), ..., f_n(s_i, a_i) \rangle = v_i$;
2: Compute clusters $\{C_1, ..., C_N\}$ of vectors $v_i$ using hierarchical clustering;
3: Compute Bayesian heuristic values of the clusters $V(C_i), i \in [N]$;
4: Identify the subset of representative clusters $\mathcal{C} = \{C_i : \frac{|C_i|}{|\mathcal{D}|} \geq X, i \in [N], X \in [0,1]\}$;
5: **repeat**
6:   Set $\hat{\mathcal{D}} = \{(s_i, a_i) : \exists C_j \in \mathcal{C} : (s_i, a_i) \in C_j\}$;
7:   Set negative demos as $\hat{\mathcal{D}}^- = \{(s_i, a') \mid \pi(a' \mid s_i) < \arg\max_{a \in \mathcal{A}} \pi(a \mid s), (s_i, a) \in \hat{\mathcal{D}}\}$;
8:   Initialize a set of candidate formulas $F$;
9:   **for** $\rho$ in $1, \ldots, d$ **do**
10:     Compute $f = LPP(\hat{\mathcal{D}}, \hat{\mathcal{D}}^-)$ with the maximum size $\rho$
        and $\hat{\mathcal{D}}, \hat{\mathcal{D}}^-$ split into train and validation sets according to the clusters;
11:     Compute the number of distinct predicates used by $f$ and denote it $p_f$;
12:     Generate $L$ rollouts for $f$ and compute its mean reward $m_f$;
13:     $F \leftarrow F \cup \{f\}$;
14:   **end for**
15:   $F \leftarrow \{f \in F : \forall f' \in F : \frac{m_{f'}}{m} < \frac{m_f}{m} + \delta\}$;
16:   Choose $f_{best} = \arg\min_{f \in F} p_f$;
17:   **if** $m_{f_{best}}/m \geq \alpha$ **then**
18:     **return** Formula simple formula $f_{best}$ which obtains a similar reward to $\pi$
19:   **end if**
20:   $C_{min} = \arg\min_{C \in \mathcal{C}} V(C)$;
21:   $\mathcal{C} \leftarrow \mathcal{C} \setminus \{C_{min}\}$;
22: **until** $\mathcal{C} = \emptyset$
23: **return** No solution for the considered set of predicates

---

Figure 3 depicts a diagram with the workflow of AI-Interpret. The computation starts with a set of demonstrations, and a domain specific language of predicates that describe the environment under consideration. The algorithm turns each of the demonstrations into a binary vector with one entry for each predicate, and with this data uses LPP to find a maximum posterior DNF formula that best explains the demonstrations. Contrary to the vanilla LPP, however, it does not stop after an unsuccessful attempt at interpretation that finds no solution or a solution that does not meet the input constraints (specified by $d$ and $\alpha$, among others). Instead, it searches for a subset of demonstrations that can be described by an appropriate interpretable decision rule. Concretely, AI-Interpret clusters the binary vectors into *J* separate sets and simultaneously assigns each a heuristic value. Intuitively, this value describes how simple it is to incorporate the demonstrations of that cluster into the final interpretable description. It then successively removes the clusters with the lowest values until LPP finds a MAP formula that is consistent with all of the remaining demonstrations and abides by the specification provided by the constraints. In this way, AI-Interpret finds a description *f* tihat belongs to $F_\alpha^{d,\mathcal{L}}$ and is nearly optimal for the maximization of $I(\cdot)$.

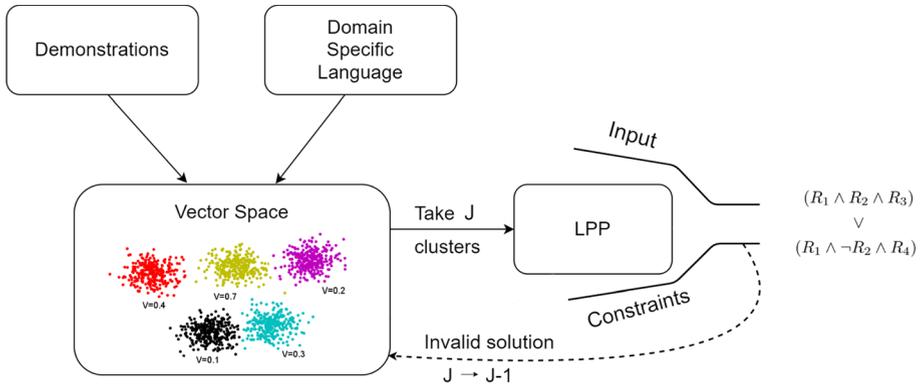

**Fig. 3** Flowchart of the AI-Interpret algorithm. Demonstrations are turned into feature vectors using the Domain Specific Language of predicates, and then clustered into sets encompassing some type of the policy's behavior. The clusters are then ordered based on how interpretable they are. The LPP method tries to iteratively construct a logical formula that imitates the policy on the demonstrations and meets the input criteria. After every failed iteration, AI-Interpret removes the least interpretable cluster and the process repeats

It is important to note that the algorithm we propose can be combined with any imitation learning method that can measure the quality of the clusters' descriptions. As such, AI-Interpret may be viewed as a general approach to finding simple policies that successfully imitate the essence of potentially incoherent set of demonstrations and to solving optimization problems similar to this in Eq. 6. In this way, AI-Interpret is able of handling sets that include some idiosyncratic or overly complicated demonstrations which do not require capturing. Moreover, by preforming imitation learning AI-Interpret can be applied to all kinds of RL policies, e.g. flat, hierarchical, metalevel, meta, etc.

### 5.3 Adaptive imitation-interpretation

In this section we describe the algorithm in more detail. Firstly, AI-Interpret accepts a set of parameters that affect the final quality and interpretability of the result. Secondly, it takes four important steps (see steps 2, 3, 7 and 9) that need to be elaborated on.

We begin with a short explication of the parameters. Note that as it was stated in Sect. 2.7, a logical formula *f* induces a policy $\pi_f$ which assumes a uniform distribution over all the actions *a* accepted by *f* in state *s*, that is $\pi_f(a \mid s) = \frac{1}{|\{a : f(s,a)=1\}|}$. While describing the parameters, we will refer to this policy as the interpretable policy.

*Aspiration value* $\alpha$ Parameter $\alpha$ specifies a threshold on the expected return ratio. For an interpretable policy $\pi_f$ to be accepted as a solution the ratio between its estimate of the expected return and the estimated expected return of the demonstrated policy has to be at least $\alpha$–$\delta$ (see the *tolerance* parameter below).

*Number of rollouts L* Parameter $L$ is a case-dependent parameter that specifies how many times to initialize a new state and run the interpretable policy in order to reliably estimate its expected return $\mathbb{E}(G_0^{\pi_f})$ (within the bounds specified by the *tolerance* parameter, see below). $L$ should be chosen according to the problem.

*Tolerance $\delta$* The parameter $\delta$ allows the user to express how much better a more complex decision rule would have to perform than a simpler rule to be preferable. Formally, the return ratio $r_2$ of the simplified strategy is considered to be significantly better than the return ratio $r_1$ of another strategy if $r_2 - r_1 > \delta$.

*Mean reward m* The mean reward of policy $\pi$ is what the interpretable policy's return tries to match in expectation. The maximum deviation from $m$ is controlled by the aspiration value, whereas the expected return of the interpretable policy is calculated by performing rollouts. Mean reward $m \approx \mathbb{E}(G_0^\pi)$.

*Maximum size d* Parameter $d$ sets an upper bound on the size of conjunctions in formulas returned by AI-Interpret. In equivalent terms, the tree that is a graphical representation of the algorithm's output is required to have the depth of at most $d$ nodes. The size of the conjunctions (or depth of the tree) is a proxy for interpretability. Decreasing the depth parameter $d$ can force the solution to use fewer predicates; this can make the formula less accurate but more interpretable. Increasing the depth may allow the method to use more predicates; this could result in overfitting and a decline of interpretability. Alongside the aspiration value and given a DSL, maximum size defines $F_\alpha^{d,\mathcal{L}}$ from our formal definition of the problem.

*Number of clusters N* The number of clusters $N$ determines how coarsely to divide the demonstrations in $\mathcal{D}$ based on the similarity of their predicate values. A proper division enables selecting a subset $\mathcal{D}_{sub} \subseteq \mathcal{D}$ such that $\mathcal{D}_{sub}$ guarantees a high probability of being captured with existing predicates, and lowers the chance of the validation set $\mathcal{D}_2 \subset \mathcal{D}_{sub}$ being largely different from $\mathcal{D}_1 \subset \mathcal{D}_{sub}$, $\mathcal{D}_1 \cap \mathcal{D}_2 = \emptyset$.

*Cut-size for the clusters X* If a cluster contains less than $X\%$ of the demonstrations, then AI-Interpret will disregard it (see step 4). Choosing representative clusters allows to remove the outliers. Since $X$ could be in fact kept fixed for virtually any problem, it is treated as a hyperparameter and does not constitute the input to the algorithm.

*Train and formula-validation set split S* Another hyperparameter of our method defines how to divide the set of demonstrations $\mathcal{D}$ to find a formula using one subset and compute its likelihood using the second subset. The split is applied to each cluster separately. Similarly to the cut-size, it could be kept fixed irrespective of the problem under consideration and hence does not constitute the input to AI-Interpret. Splitting is performed in each iteration by randomly dividing the clusters into train and validation sets and then using the sum of all train sets and the sum of all validation sets as inputs to LPP.

Now we move to the explanation of the steps taken by the algorithm and start with the isolated case of step 7. In the original formulation of LPP, the authors take all the actions that were not taken in a demonstration $(s, a)$ to serve as the negative examples $(s, a')$, $a' \neq a$. Since in our problem we do have access to the policy, we use a more conservative method and select only the state-action pairs which are sub-optimal with respect to $\pi$. This helps the algorithm find more accurate solutions.

In step 2, the algorithm uses feature vectors corresponding to predicate values and clusters them into separate subsets. It is done through hierarchical clustering with the UPGMA method (Michener & Sokal, 1957), as this method captures the intuition that there may exists a core of predicates which evaluate to the same value for the demonstrations forming the cluster, and that there might also exist irrelevant predicates making up the noise. The elements of the cluster identified by hierarchical clustering are hence well poised to capture different sub-behaviors of the demonstrated policy.

With step 3 the algorithm measures which of the clusters are indeed well described with the predicates. The Bayesian heuristic value of a cluster (Definition 6) is defined as the MAP estimate of its interpretable description found by Logical Program Policies, weighted by the size of the cluster relative to the size of the whole set. The larger the value, the more similar behavior is encompassed by the elements of the cluster, and the more representative it is. Note, that through step 3 (and after applying the cut-size in step 4) it becomes possible to rank order the clusters. In case of a failure in interpreting the policy with existing examples, the cluster with the lowest rank can be removed—see step 21. In this way, the algorithm may disregard a set of demonstrations that are not described with existing predicates as well as the others, and continue with the remaining ones.

In step 9, our algorithm uses the LPP method to extract formulas of progressively larger conjunctions, up to size $d$ specified by the user. It then selects formulas which are not significantly worse than other found ones (according to the tolerance parameter, see step 15), and eventually chooses the formula with the fewest predicates (step 16). This allows our algorithm to consider all decision rules that could be generated for the same (incomplete) demonstration dataset, and return the best and the simplest among them.

The solution is output as soon as the expected reward of the interpreted policy is close enough to the expected reward of the original policy (step 17). If that never happens, the algorithm concludes that the set of predicates is insufficient to satisfy the input constraints.

**Definition 6** (*Bayesian heuristic value*) For a subset $\mathcal{C} \subseteq \mathcal{D}$ extracted from a dataset of demonstrations $\mathcal{D}$ the Bayesian heuristic value of this set is given by:

$$V(\mathcal{C}) = P(LPP(\mathcal{C}) \mid \mathcal{C}) \frac{|\mathcal{C}|}{|\mathcal{D}|}. \tag{7}$$

### 5.4 Choosing the number of clusters

In this section, we introduce a heuristic that helps to narrow down the list of candidates for parameter $N$ of the algorithm, that is the number of clusters.

In more detail, we adapt the popular elbow heuristic. Our version of this heuristic (see Procedure 1) allows to choose a subset of values for the number of clusters, by specifying how fine-grained the clustering is required to be to most drastically change the Clustering Value (see Definition 7). The Clustering Value of $N$, $CV(N, X)$, is the sum of Bayesian heuristic values (Definition 6) of all the clusters found by hierarchical clustering with size is at least $X$% of the whole set. Practically, we use the same $X$ that serves as the cut-size hyperparameter for AI-Interpret, see Sect. 5.3. A leap in the values of $CV$ conveys that the clusters are relatively big and much better described in terms of the predicates as they were for a coarser clustering. We search for an elbow in the Clustering Values because we would like the clusters to be maximally distinct while keeping their number as small as possible. Finally, the heuristic returns a set of candidate elbows since *a priori* the granularity of the data revealed by the predicates is unknown.

**Definition 7** (*Clustering value*) We define the Clustering Value function $CV : \mathbb{N} \times [0, 1] \to \mathbb{R}$ as

$$CV(N, X) = \sum_{C \in \{C_i : |C_i|/|\mathcal{D}| \geq X, i \in [N]\}} V(C) \qquad (8)$$

$$= \sum_{C \in \{C_i : |C_i|/|\mathcal{D}| \geq X, i \in [N]\}} P(LPP(C) \mid C) \frac{|C|}{|\mathcal{D}|}. \qquad (9)$$

where $V$ stands for the Bayesian heuristic value function, $N$ denotes the number of clusters $C_1, \ldots, C_N$ identified by hierarchical clustering on dataset $\mathcal{D}$, that is $\bigcup_{i=1}^{N} C_i = \mathcal{D}$ and $C_i \cap C_j = \emptyset$ for $i \neq j$, and $X$ stands for the cut-size value for the size of clusters measured proportionally to $\mathcal{D}$.

**Procedure 1** (Elbow heuristic) *To decide the number of clusters, fix the cut-size hyperparameter $X$ and evaluate the Clustering Value function CV on the set of m candidate values $N_1, \ldots, N_m$. Identify K values $N_{i_1}, \ldots, N_{i_K}$ for which $CV(N_{i_j}, X) - CV(N_{i_{j-1}}, X)$, $j \leq K$ is the largest, i.e. identify the elbows. The elbows heuristically identify clustering solutions for which the elements within each cluster are similar to one another, can be appropriately described by the predicates, and convey that clusters are reasonably large.*

If the value does not ever increase substantially, then the predicates do not capture the general structure of the data. An example of how to use the elbow heuristic is shown in Fig. 4.

### 5.5 Pipeline for interpretable strategy discovery

To go from a problem statement to an interpretable description of a strategy that solves that problem we take three main steps: (1) formulate the problem in formal terms, (2) discover the optimal strategy using this formulation, (3) interpret the discovered strategy. The first two steps can be broken down to sub-problems that our previous work has already solved (Callaway et al., 2018a, 2018b, 2019) and which we have described in Sect. 3. The last step is feasible through AI-Interpret. We show a pseudo-code which implements the pipeline for automatic discovery of interpretable planning strategies in Algorithm 2.

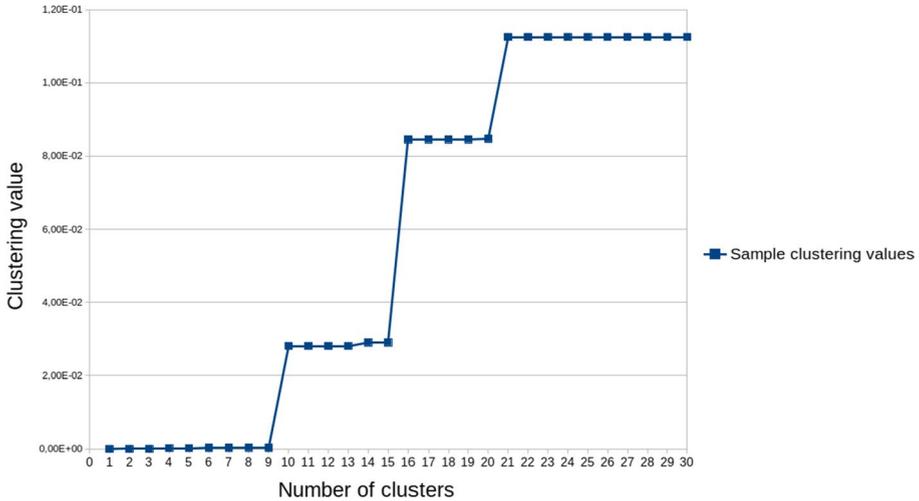

**Fig. 4** Sample plot of the Clustering Value function. For $K = 3$, the candidates would in this case comprise 10, 16 and 21 clusters

---

**Algorithm 2:** Automatic Discovery of Interpretable Planning Strategies

**Input:** Problem $p$;
**Output:** Set of interpretable descriptions $DT(p)$.
1: Model the decision problem $p$ and the decision maker's cognitive architecture as a metalevel MDP.
2: Solve the metalevel MDP with dynamic programming or metalevel RL to obtain a resource-rational policy $\pi^\star_{\text{meta}}$.
3: Extract the expected reward $m$ of policy $\pi^\star_{\text{meta}}$ as $\max_{c \in \mathcal{C}} Q^*_{\text{meta}}(b_{init}, c)$.
4: Perform $x$ rollouts of the policy $\pi^\star_{\text{meta}}$ in its metalevel MDP, saving $n$ obtained demonstrations in $\mathcal{D}_{meta} = \{(b_i, c_i) : i \in [n]\}$.
5: Create a Domain Specific Language $\mathcal{L}$ that contains predicates describing the characteristics of the real environments.
6: Choose the maximum values for parameters for AI-Interpret: $\delta = \Delta, \alpha = A, d = D$, and choose a sufficient number of rollouts $L$.
7: Fix common cut-size hyperparameter for the clusters' sizes $X$ and split for the demonstration set $S$.
8: Choose $K$ and use the elbow heuristic to determine $K$ candidates $N_1, \cdots, N_K$ defining how many clusters to use to separate the demonstrations.
9: **for** $i \in 1, \cdots, K$ **do**
10:  $f_i = \text{AI\_Interpret}(\mathcal{D}_{meta}, \mathcal{L}, L, \Delta, m, A, D, N_i)$.
11:  Turn $f_i$ into a graphical representation of a decision tree $dt(f_i)$.
12:  $DT(p) \leftarrow DT(p) \cup \{(f_i, dt(f_i))\}$;
13: **end for**
14: **return** $DT(p)$

---

Our pipeline starts with modeling the problem as a metalevel MDP and then solving it to obtain the optimal policy. To use AI-Interpret, the found policy is used to generate a set of demonstrations. We also create a DSL of predicates that is used to provide an interpretable description of this policy. We then establish the input to AI-Interpret. The mean reward of the metalevel policy is extracted directly from its Q function taking the maximum at the initial state. The number of clusters $N$ is identified automatically by the elbow heuristic (Procedure 1). Since the elbow heuristic returns $K$ candidates for $N$, our pipeline outputs a set of $K$ possible interpretable descriptions, each for a different clustering. Having a dataset of candidate interpretable descriptions (decision trees) output by the pipeline, one may use background knowledge or a pre-specified criterion to choose the most interpretable tree. Criteria include, but are not limited to, choosing the tree with the least amount of nodes, the interpretability ratings of human judges, or the performance of people who are assisted by alternative decision trees. Our method of extracting the final result is detailed in Sect. 6.2.

# 6 Improving human decision-making

Having developed a computational pipeline for discovering high-performing and easily-comprehensible decision rules, we now evaluate whether this approach meets our criteria for interpretability. As we mentioned in Sect. 3, our evaluation introduces a new standard to the field of interpretable RL by studying how the descriptions affect human decision makers. As a proof-of-concept, we test our approach on three types of planning tasks that are challenging for people (Callaway et al., 2018b; Lieder et al., 2020). The central question is whether we can support decision-makers in the process of planning by providing them with flowcharts discovered through our computational pipeline for interpretable strategy discovery (see Fig. 1). To achieve that, we perform one large behavioral experiment for each of the three types of tasks and a fourth experiment in which we evaluate our approach to improving human decision-making against conventional training. We find that our approach allows people to largely understand the automatically discovered strategies and use them to make better decisions. Importantly, we also find that our new approach to improving human decision-making is more effective than conventional training.

## 6.1 Planning problems

Human planning can be understood as a series of information gathering operations that update the person's beliefs about the relative goodness of alternative courses of actions (Callaway et al., 2020). Planning a road trip, for instance, involves gathering information about the locations one might visit, estimating the value of alternative trips, and deciding when to stop planning and execute the best plan found so far. The Mouselab-MDP paradigm (Callaway et al., 2017) is a computer-based task that emulates these kinds of planning problems (see Fig. 5). It asks people to choose between multiple different paths, each of which involves a series of steps. To choose between those paths, people can gather information about how much reward they will receive for visiting alternative locations by clicking on the corresponding location. Since people's time is valuable, gathering this information is costly (each click costs $1) but it can also improve the quality of their decisions. Therefore, a good planning strategy has to focus the decision-maker's attentions on the most valuable pieces of information.

To test our approach to improving human decision-making, we rely on three route-planning tasks that were designed to capture important aspects of why it is difficult for people to make good decisions in real-life (Callaway et al., 2018b). For instance, the first task captures that certain steps that are very valuable in the long-run (e.g., filing

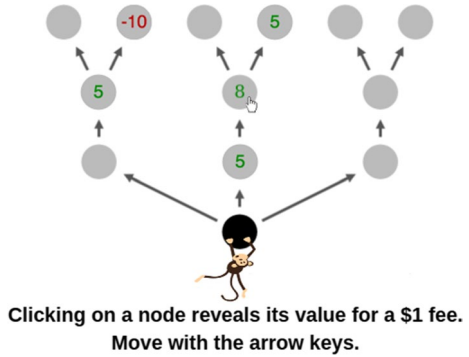

**Fig. 5** The experimental task: click the fewest nodes possible and help the monkey to climb up a tree through a path with the highest possible rewards

taxes) are often unrewarding in the short-run whereas activities that are rewarding in the short-run (e.g., watching cat videos on YouTube) often have little value in the long-run. The three tasks have been previously used to study how people plan (Callaway et al., 2017, 2018b), to train people how to make more far-sighted decisions (Lieder et al., 2019, 2020), and to compare the effectiveness of different ways to improve human decision-making (Lieder et al., 2020).

The route planning problems we presented to our participants used the tree environment illustrated in Fig. 5. The node at the bottom of this tree served as the starting node and was connected to 3 other nodes, each of those was connected to one additional node, and finally each of these single connections led to 2 further nodes. We will call this a 3-1-2 structure and refer to the nodes that can be reached in 1, 2, and 3 steps as level 1, level 2, and level 3, respectively. Callaway et al. (2018b) defined the three environments we are using in terms of the distribution of rewards for nodes on each level. These environments differ in that the uncertainty about the rewards either increases, decreases, or stays the same from each step to the next. They created three discrete sequences of discrete uniform distributions, further called variance structures. Building on these structures, we defined three types of environments. Within each type of an environment, the rewards of all nodes at the same level were drawn from the same discrete uniform distribution. As shown below, what distinguishes the environments is the assignment of reward distributions to levels (read as *level* : *support*):

1. *Increasing variance* structure environments where
   $1 : \{-4, -2, 2, 4\}, \ 2 : \{-8, -4, 4, 8\}, \ 3 : \{-48, -24, 24, 48\}$
2. *Decreasing variance* structure environments where
   $1 : \{-48, -24, 24, 48\}, \ 2 : \{-8, -4, 4, 8\}, \ 3 : \{-4, -2, 2, 4\}$
3. *Constant variance* structure environments where
   $1 : \{-10, -5, 5, 10\}, \ 2 : \{-10, -5, 5, 10\}, \ 3 : \{-10, -5, 5, 10\}$

Discovering a resource-rational planning strategy corresponds to computing the optimal policy of a metalevel MDP that is exponentially more complex that the acting domain itself. In fact, the number of belief states for all our planning problems equals $4^{12}$, that is over 16 million. This shows that optimizing a policy for Mouselab-MDP is in reality a very difficult problem. Moreover, creating a description of the optimal policy also poses scalability issues. In particular, the standard symbolic approaches that derive rules from the Q-table of the policy would result in overwhelmingly large set of rules if they accounted for even a fraction of the original metalevel belief state space.

Prior research on the constant and increasing variance environments indicates that tasks in the Mouselab-MDP paradigm come as a challenge to many people (Callaway et al., 2018b; Lieder et al., 2020). In the case of the constant variance structure, even extensive practice is not sufficient for participants to arrive at nearly-optimal strategies (Callaway et al., 2018b). We aimed to help people adopt good approximations to those strategies by the virtue of showing them a decision aid for playing a game defined in the world of Mouselab MDPs. To evaluate our method's potential for helping people make better decisions in this game, we designed a series of online experiments with real people. In each, participants were making decisions with versus without the support of an interpretable flowchart. In the game we were studying, participants helped a monkey to climb up a tree through a path that enables it to get the highest possible reward, see Fig. 5. Their decisions regarded uncovering the hidden nodes in search of such paths, knowing that each uncovered reward takes some money away from the monkey ($1).

### 6.2 Designing decision aids with AI-interpret and automatic strategy discovery

To apply our interpretable strategy discovery pipeline (see Fig. 1) to the benchmark problems described in Sect. 6.1, we modeled the optimal planning strategy for each of the three types of sequential decision problems as the solution to a metalevel MDP (see Definition 2) as it was previously done by Callaway et al. (2018b). The belief state of the metalevel MDP corresponds to which rewards have been observed at which locations. The computational actions of the metalevel MDP correspond to the clicks people make to reveal additional information. The cost of those computations is the fee that participants are charged for inspecting a node. We obtained the optimal metalevel policies for the three metalevel MDPs using the simplest approach we had at our disposal, that is dynamic programming method developed by Callaway et al. (2018b). It was distinct from the original algorithm by being able to handle much larger state space typical for metalevel MDPs.

Afterwards, we employed our pipeline for automatic discovery of interpretable strategies from Algorithm 2. First, we generated three sets of 64 demonstrations by running the optimal metalevel policies on their respective metalevel MDPs. Then, we established a novel domain specific language (DSL) $\mathcal{L}$ to allow AI-Interpret to describe the demonstrated planning policies in logical sentences via LPP. Our DSL $\mathcal{L}$ supplied LPP with the basic building blocks ("words") for describing the Mouselab environment and the demonstrated information gathering operations. In more detail, $\mathcal{L}$ comprised six types of predicates:

- *PRED*($b$, $c$) describes the node evaluated by computation $c$ in belief state $b$,
- *AMONG*(*PREDS*($b$, $c$)) checks if the node evaluated by $c$ in state $b$ satisfies the predicate or a conjunction of predicates *PREDS* of the type as above,
- *AMONG_PRED*($b$, $c$, *list*) checks if the node evaluated by $c$ in state $b$ belongs to *list* and possesses a special characteristic as the only one in that set (not used on its own),
- *AMONG*(*PREDS*, *AMONG_PRED*)($b$, $c$) checks if in belief state $b$ the node evaluated by $c$ satisfies the predicate *AMONG_PRED* among the nodes satisfying the predicate or a conjunction of predicates *PREDS*,
- *ALL*(*PREDS*, *AMONG_PRED*)($b$, $c$) checks if in belief state $b$ all the nodes which satisfy the predicate(s) *PREDS* also satisfy *AMONG_PRED*,

– *GENERAL_PRED*(*b*, *c*) detects whether some feature is present in belief state *b*.

A probabilistic context-free grammar with 14 base *PRED* predicates, 15 *GENERAL_PRED* predicates, and 12 *AMONG_PRED* predicates generated $\mathcal{L}$ according to the above-mentioned types. This resulted in a set containing a total of 14,206 elements. More information on our DSL can be found in the Supplementary Material. The predicates found in the flowcharts we used for the benchmark problems were of the following types:

*GENERAL_PRED*, *AMONG*(*PREDS*) and *AMONG*(*PREDS*, *AMONG_PRED*).

Other parameters necessary to employ AI-Interpret in the search of interpretable descriptions comprised number of rollouts, aspiration value, tolerance, maximum size, number of clusters, mean reward of the expert policy. Preliminary runs performed to establish the expected return of the optimal metalevel policy or of policies which behaved similarly, revealed that $L = 100,000$ is the number of rollouts appropriate for all the studied problems. The aspiration value $\alpha$ was fixed at 0.7 and the tolerance parameter $\delta$ was equal 0.025. The maximum depth $d$ had a limit value of 5 and the number of clusters $N$ was chosen by the elbow heuristic employed in Algorithm 2. Eventually, $N$ was set to 18 for the increasing and constant variance environments and set to 23 for the decreasing variance environment. Clusters were created based on the output of the UPGMA hierarchical clustering with $l_1$ distance and average linkage function (Michener and Sokal, 1957). We extracted the mean rewards of the optimal policies by inspecting their Q function in the initial belief state $b_0$. We also used 2 hyperparameters. To reject the outliers, both in the elbow heuristic and the algorithm, any cluster whose size was less than $X = 2.5\%$ of the whole set of demonstrations was disregarded. The split $S$ for the demonstration set to validate the formulas in each iteration was equal to 0.7.

Applying AI-Interpret with this DSL and parameters to the demonstrations induced the formulas that were most likely to have generated the selected demonstrations. These formulas were subject to inductive constraints of our DSL and the simplicity required by the listed parameters. The formal output of our pipeline (Algorithm 2) comprised a set of $K = 4$ decision trees defined in terms of logical predicates. We chose one output per decision problem by selecting the tree with the fewest nodes, breaking ties in favor of the decision tree with the lowest depth. To obtain fully comprehensible decision aids we turned those decision trees into human-interpretable flowcharts by manually translating their logical predicates into natural language questions. A translation of the *GENERAL_PRED* predicates depended on the particular characteristic they were capturing. For example, a predicate `is_previous_observed_max` was translated as "Was the previously observed value a 48?". The translation of *AMONG* predicates was constructed based on the following prototype: "Is this node/it PRED AMONG_PRED", "Is this node/it PRED and PRED" or "Is this node/it AMONG_PRED among PRED nodes". For instance, `among(not(is_observed), has_largest_depth)` was translated as "Is it on the highest level among unobserved nodes?". The flowcharts led to two possible high-level decisions, which we named "Click it" and "Don't click it". The termination decision was reached when all the possible actions led to "Don't click it" decision. The flowcharts were eventually polished in the pilot experiments by asking participants for their semantic preferences and by incorporating comments that they submitted.

By applying this procedure to the three types of sequential decision problems described above, we obtained the three flowcharts shown in Fig. 6. The flowchart for the increasing variance environment (Fig. 6a), advises people to inspect nodes of the third level until they uncover the best possible reward (+48) and then suggests to stop planning and take action. Despite being simpler, this strategy performs almost as well as the optimal metalevel

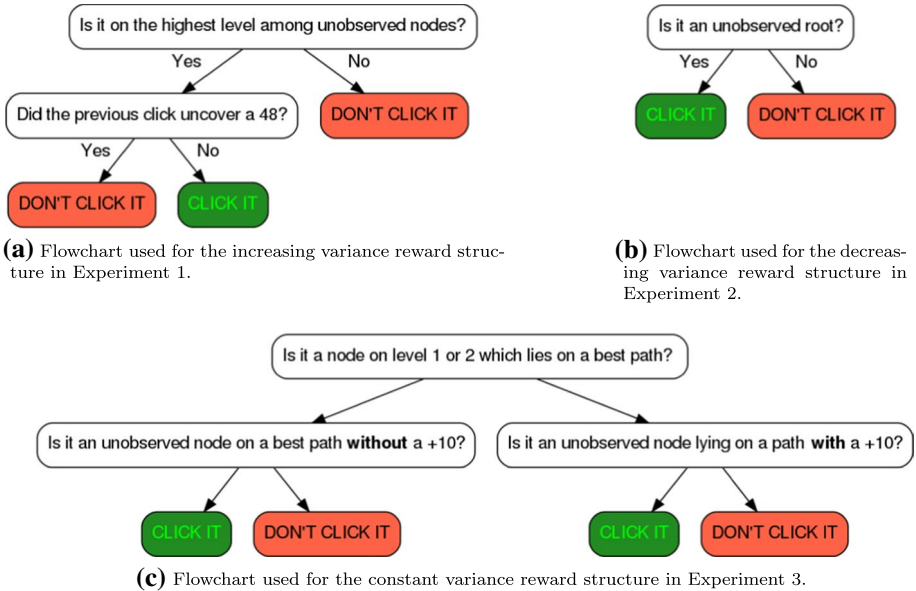

**(a)** Flowchart used for the increasing variance reward structure in Experiment 1.

**(b)** Flowchart used for the decreasing variance reward structure in Experiment 2.

**(c)** Flowchart used for the constant variance reward structure in Experiment 3.

**Fig. 6** Interpretable flowcharts generated by applying the procedure described in this subsection and shown to people in the experiments on improving human decision-making

policy that it imitates (39.17 points/episode vs. 39.97 points/episode). The flowchart for the decreasing variance environment (Fig. 6b) allows people to move after clicking all level 1 nodes. This strategy is also simpler that the optimal metalevel policy that it imitates, and performs almost as well (28.47 points/episode vs. 30.14 points/episode). The flowchart for the constant variance environment (Fig. 6c) instructs a person to click level 1 or 2 nodes that lie on the path with the highest expected return until observing a reward of +10. Then, it suggests to click the two level 3 nodes that are above the +10, and then either get back to clicking on level 1 or 2 or, if the best path is a path that passes through the +10, stop planning and take action. In this case we also note that this strategy is simpler than the optimal metalevel policy that it imitates and that it once again performs almost as well (7.03 points/episode vs. 9.33 points/episode).

### 6.3 Evaluation in behavioral experiments

We evaluated the interpretability of the decision aids designed with the help of automatic strategy discovery and AI-Interpret in a series of 4 behavioral experiments. In the first three experiments we evaluated whether the flowcharts generated by our approach were able to improve human decision-making in the Mouselab-MDP environments with increasing variance, decreasing variance, and constant variance, respectively. In the last experiment we investigated for which environments our approach is more effective than the standard educational method of giving people feedback on their actions.

In each experiment participants were posed a series of sequential decision problems using the interface illustrated in Fig. 5. In each round, the participant's task was to collect the highest possible sum of rewards, further called the score, by moving a monkey up a tree along one of the six possible paths. Nodes in the tree harbored positive or negative rewards and were initially occluded; they could be made visible by clicking on them for a fee of $1 or moving on top of them after planning. The participant's score was the sum of the rewards along their chosen path minus the fees they had paid to collect information.

Our experiments focused on two outcome measures: the expected score and the click agreement. The expected score is the sum of revealed rewards on the most promising path in the round right before the participant started to move, minus the cost of his or her clicks. This is true because the expected reward of an occluded node was 0 in all of the chosen decision problems. The fact that the expected score is equal to the value of the termination operation of the corresponding metalevel MDP, makes it the most principled performance metric we could choose. It is also the most reliable measure of the participant's decision quality because it is their expected performance across all possible environments that are consistent with the observed information. The total score, by contrast, includes additional noise due to the rewards underneath unobserved nodes. Our second outcome measure, the click agreement, quantifies a person's understanding of the conveyed strategy by measuring how many of his or her clicks or consistent versus inconsistent with that strategy. When the participant clicked a node for which the flowchart said "Click it" this was considered a consistent click. When the participant clicked a node that the flowchart evaluated as "Don't click it" this was considered as an inconsistent click. We defined the click agreement as the proportion of consistent clicks in relation to all performed clicks, that is

$$\text{agreement} = \frac{n_{\text{consistent}}}{n_{\text{consistent}} + n_{\text{inconsistent}}}.$$

When people made fewer clicks than the flowchart suggested, then the difference between the number of clicks made by the strategy shown in the flowchart and the participant's number of clicks was counted towards the number of inconsistent clicks. The number of clicks made by the strategy was estimated by its average number of clicks across 1000 simulations.

Differences in the click agreement between participants belonging to separate experimental groups are indicative of differences in understanding the strategy. If one group has significantly higher click agreement, it means that people belonging to that group know the strategy better than people in the other group(s). Our goal was to show that the click agreement for participants who were assisted with our flowcharts was significantly higher than the click agreement for participants not using such an assistance. This would not only show that flowcharts are interpretable, but it would also rule out the possibility that people already knew the strategy or were able to discover it themselves. Since the flowcharts our method finds convey near-optimal strategies, higher click agreement is correlated with higher performance. However, it is unclear what level of understanding is sufficient for participants to benefit from the strategy. To show that the flowcharts are not only interpretable, but interpretable enough to increase people's performance, we additionally measured our participants' expected score.

### 6.3.1 Experiment 1: improving planning in environments with increasing variance

In the first experiment, we evaluated whether the flowchart presented in Fig. 6a can improve people's performance in sequential decision problems where the uncertainty about the reward increases from each step to the next.

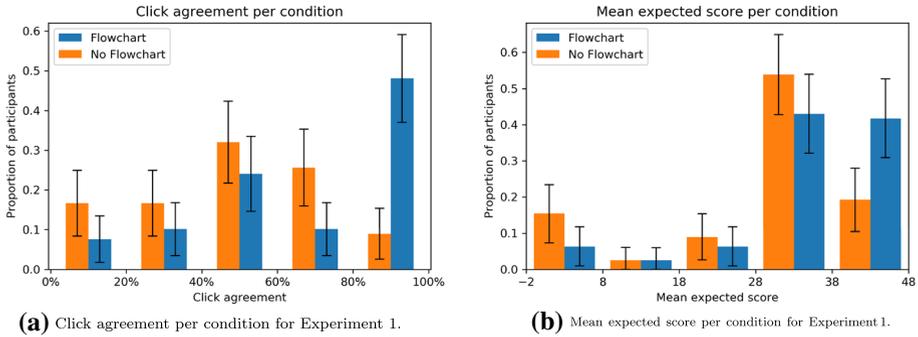

(a) Click agreement per condition for Experiment 1.

(b) Mean expected score per condition for Experiment 1.

**Fig. 7** Automatic discovered flowchart for the increasing variance environment is interpretable and improves planning: **a** Proportion of people arranged in five equal sized bins based on the click agreement. **b** Proportion of people arranged in five equal sized bins based on the average expected score. The error bars enclose 95% confidence intervals. Orange bars represent the control group and blue bars represent the experimental group in Experiment 1

*Procedure* Each participant was randomly assigned to either a control condition in which no strategy was taught or an experimental condition in which a flowchart conveyed the strategy. The control condition consisted of an introduction to the Mouselab-MDP paradigm including three exploration trials, a quiz to test the understanding of the instructions, ten test trials, and a short survey. We instructed participants to maximize their score using their own strategy and incentivized them by communicating to pay an undefined score-based bonus. After the experiment, they received 2 cents for each virtual dollar they had earned in the game. The experimental condition consisted of an introduction to the Mouselab-MDP paradigm including one exploration trial, an introduction to flowcharts and their terminology including two practice trials, a quiz to test the understanding of the instructions, ten test trials, and a short survey. The practice and test trials displayed a flowchart next to the path-planning problem. The flowchart used in the practice block did not convey a reward enhancing strategy to avoid a training effect, whereas the one in the test block did. We instructed participants to act according to the displayed flowchart. Specifically, participants were asked to first click all nodes for which following the flowchart led to the "Click it" decision and to then move the agent (a monkey) along the path with the largest sum of revealed rewards. To incentivize participants, they were told to receive a bonus depending on how well they followed the flowchart and they received a bonus that was proportional to their click agreement score.

*Participants* We recruited 172 people on Amazon Mechanical Turk (average age: 37.9 years, range: 18–69 years; 85 female). Each participant received a compensation of $0.15 plus a performance based bonus of up to $0.65. The mean duration of the experiment was 10.3 min. On average, participants needed 1.4 attempts to pass the quiz. Because not clicking is highly indicative of speeding through the experiment without engaging with the task, we excluded 15 participants (i.e., 8.72%) who did not perform any click in the test block. This yielded 78 participants for the control condition and 79 participants for the experimental condition.

*Results* In addition to a significant Shapiro-Wilk test for normality ($W = .94, p < .001$), we observed that the distribution of click agreements is highly left-skewed. Due to this reason we report the median values instead of the mean values for the click agreement and use the non-parametric Mann–Whitney U-test for statistical comparisons. The median click agreement was 46.6% ($M = 46.9\%, SD = 25.8\%$) in the control condition and 74.1% ($M = 67.0\%, SD = 28.2\%$) in the experimental condition (see Fig. 7a). The proportion of people who achieved the click agreement above 80% increased from 9% without the flowchart to 48% with the flowchart. Similarly, the proportion of participants who achieved the click agreement above 50% increased from 46 to 70%. A two-sided Mann–Whitney U-test revealed that the click agreements in the experimental condition were significantly higher than in the control condition ($U = 1773.5$, $p < .001$). Thus, participants confronted with the flowchart followed its intended strategy more often than participants without the flowchart.

The distribution of the mean score was non-normal due to the Shapiro-Wilk test ($W = .77, p < .001$), hence we used the median values and the Mann–Whitney U-test again. The median expected score per trial in the control condition was 33.25 ($M = 28.41, SD = 13.33$) and 36.80 ($M = 34.47, SD = 11.85$) in the experimental condition. This corresponds to 83.2% and 92.1% of the score of the optimal strategy, respectively. A two-sided test showed that the expected scores in the experimental condition were significantly higher than in the control condition ($U = 2110$, $p < .001$). Thus, the flowchart positively affected people's planning strategies as participants assisted by the flowchart revealed more promising paths before moving than participants who acted at their own discretion.

In total, participants both understood the strategy conveyed by the flowchart (higher click agreement) and used it, which increased their scores (higher expected rewards).

### 6.3.2 Experiment 2: improving planning in environments with decreasing variance

In the second experiment, we evaluated whether the flowchart presented in Fig. 6b can improve people's performance in environments where the uncertainty about the reward decreases from each step to the next.

*Procedure* The experimental procedure used in this study was identical to the one presented for Experiment 1 (see Sect. 6.3.1) except that the task used the decreasing variance environment instead of the increasing variance environment and that participants in the experimental condition where correspondingly shown the flowchart in Fig. 6b.

*Participants* We recruited 152 people on Amazon Mechanical Turk (average age 36 years, range: 20–65 years; 62 female). Each participant received a compensation of $0.50 plus a performance based bonus up to $0.50. The mean duration of the experiment was 8.2 min. The participants needed 1.7 attempts to pass the quiz on average. We excluded 6 participants (3.95%) who did not perform any click in the test block. This resulted in 70 participants for the control condition and 76 participants for the experimental condition.

*Results* Similarly to Experiment 1, Shapiro–Wilk tests revealed that the dependent variables in Experiment 2 were not normally distributed ($W = .94, p < .001$ for the click-agreement and $W = .98, p = .019$ for the expected scores). Hence, we report the median values and the results of the Mann–Whitney U-test here as well. The median click agreement in the control condition was 44.7% ($M = 48.6\%, SD = 19.8\%$) and 65.7% ($M = 70.5\%, SD = 24.0\%$) in the experimental condition (see Fig. 8a). The proportion of people who achieved the click agreement above 80% increased from 8.57% without the flowchart to 36.84% with the flowchart. Similarly, the proportion of participants who achieved the click agreement above 50% increased from 35.71 to 81.57%. Participants confronted with the flowchart followed its intended strategy significantly more often than participants without the flowchart ($U = 1221.0$, $p < .001$).

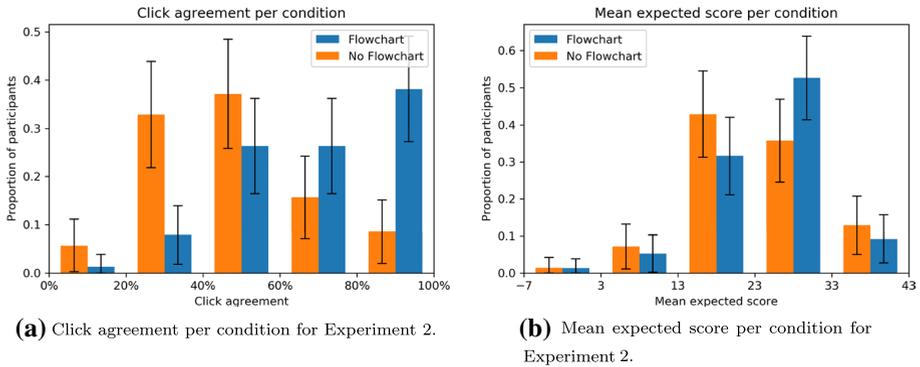

(a) Click agreement per condition for Experiment 2.

(b) Mean expected score per condition for Experiment 2.

**Fig. 8** Automatic discovered flowchart for the decreasing variance environment is interpretable: **a** Proportion of people arranged in five equal sized bins based on the click agreement. **b** Proportion of people arranged in five equal sized bins based on the average expected score. The error bars enclose 95% confidence intervals. Orange bars represent the control group and blue bars represent the experimental group in Experiment 2

The median expected score per trial was 22.85 ($M = 22.75, SD = 8.18$) in the control condition and 24.45 ($M = 24.02, SD = 7.38$) in the experimental condition. This corresponds to 75.8% and 81.1% of the score of the optimal strategy, respectively. Although the difference between the experimental condition and the control condition was not statistically significant (($U = 2384.0, p = .140$)), higher click agreement was significantly correlated with higher expected score ($r(144) = .36, p < .001$).

Thus, similarly to the previous experiment, we observed that participants did understand the strategy conveyed by the flowchart what resulted in significantly higher click agreement. Still, a small sample size and a less challenging environment (looking at immediate rewards is more intuitive than inspecting distant outcomes) prevented us from detecting a significant difference in the expected rewards.

### 6.3.3 Experiment 3: improving planning in environments with constant variance

In the third experiment, we evaluated whether the flowchart presented in Fig. 6c can improve people's performance in an environment where the uncertainty about the reward is the same in each step.

*Procedure* Since the flowchart for the constant variance environment is more complex than the flowcharts for the other two environments, participants in Experiment 3 were trained more extensively than participants in Experiments 1 and 2. The goal of this procedure was to familiarize the experimental group with the flowchart as well as possible so that they could use it during the testing phase. To minimize differences between the experimental condition and the control condition that could lead to asymmetric attrition, both groups went through the same training procedure, but only the experimental group was supported by the flowchart during the test trials.

Both conditions consisted of an introduction to the Mouselab-MDP paradigm including one exploration trial, an introduction to the terminology used in the flowcharts, a quiz and a training phase on the introduced notions, an introduction to flowcharts *per se*, a second quiz, and a practice phase on flowcharts. Each quiz consisted of 3 simple questions to check attentiveness. During training, participants answered three different questions about highlighted nodes in a partially revealed training tree. These questions had the same structure as the questions used in the testing flowchart but asked for different values, e.g. is it an unobserved node lying in a path with a −8. They were given feedback on their answers and could advance to the next question only after answering correctly. In each training round participants were sequentially quizzed about six randomly selected nodes. After the participant answered two questions about a node, the node was uncovered and the selection mark moved to another node. There were at least 3 and at most 10 training rounds. A participant was allowed to end the training after he or she had answered each of the flowchart's three questions at least 15 times and achieved an accuracy of at least 75% on each of them. The training phase was followed by an introduction to flowcharts and another quiz on understanding the task. The last block before the test phase comprised 2 practice rounds with a practice flowchart. This flowchart used only the questions presented in the training phase and, as previously, so as to minimize the effect of the shared training block on people's choices in the test block. Participants were required to first select a candidate node and sequentially answer the questions that the flowchart asked about it until the flowchart reached a decision about whether or not to click on the node. According to this decision, they were either allowed to reveal the selected node or not. Participants could not move the monkey before they had revealed all nodes that the flowchart suggested clicking. Finally, due to a large number of training trials, we used an increasing variance structure throughout the non-test trials, eliminating the possibility of implicit learning. The test block presented participants with planning problems in the constant variance environment. The experimental condition differed from the control condition only in the setup of the test block. That is, in the 10 test trials the experimental group was assisted by the flowchart whereas the control condition was not. To minimize differences in the duration of the test block, the control group completed 15 additional problems after the 10 test rounds; those additional problems were not considered in the analysis.

In contrast to the previous experiments, in Experiment 3 the flowchart was not visible as a whole during the test rounds. Rather, participants had to go through the flowchart by answering two consecutive questions interactively until they reached a decision about whether or not to click the selected node. Participants did not receive feedback on their answers, nor were they bound to the flowchart's decision. In addition, when a participant in the control condition attempted to move after having revealed less than three nodes, a dialogue informed them that "Many people overlook some of the nodes that the flowchart allows clicking and miss the bonus." and asked them "Are you sure you want to move?". The control group was promised and paid 2 cents of bonus for each dollar they scored in the game, whereas the experimental group could earn or loose 10 cents of bonus depending on whether a click was congruent with the flowchart or not.

*Participants* We recruited 149 people on Amazon Mechanical Turk (average age 33.7 years, range: 18–65 years; 62 female). Each participant received a compensation of $3 plus a performance based bonus up to $6. The mean duration of the experiment was 51 min. The participants needed two attempts on average to pass any of the quizzes. We excluded 30 participants (20.1%) who required four or more attempts on one of them. This resulted in 60 participants for the control condition and 59 participants for the experimental condition.

*Results* The Shapiro-Wilk test revealed that the click agreement in Experiment 3 was not normally distributed ($W = .95, p < .001$). Due to this fact, we report the median values and the results of the Mann–Whitney U-test for both of the dependent variables. The median click agreement was 33.5% ($M = 31.0\%, SD = 15.9\%$) in the control condition and 54.0% ($M = 59.0\%, SD = 24.4\%$) in the experimental condition (see Fig. 9a). The proportion of

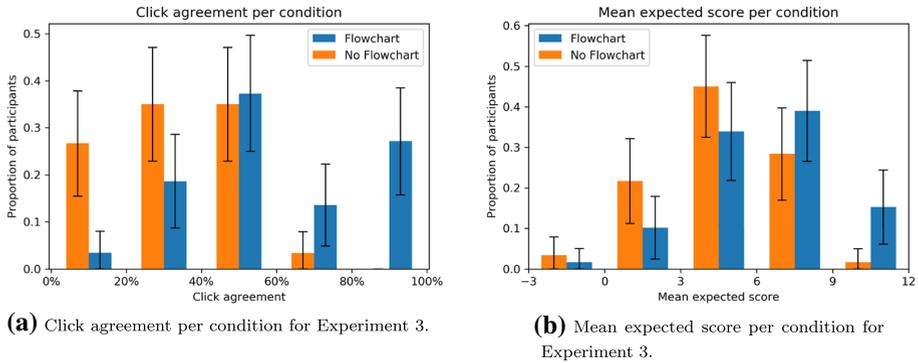

**(a)** Click agreement per condition for Experiment 3.

**(b)** Mean expected score per condition for Experiment 3.

**Fig. 9** Automatic discovered flowchart for the constant variance environment is interpretable and improves planning: **a** Proportion of people arranged in five equal sized bins based on the click agreement. **b** Proportion of people arranged in five equal sized bins based on the average expected score. The error bars enclose 95% confidence intervals. Orange bars represent the control group and blue bars represent the experimental group in Experiment 3

people who achieved the click agreement above 80% increased from 0% without the flowchart to 25.42% with the flowchart. Similarly, the proportion of participants who achieved the click agreement above 50% increased from 10 to 57.62%. Participants confronted with the flowchart followed its intended strategy significantly more often than participants without flowchart ($U = 651.0$, $p < .001$).

The median expected score per trial was 4.00 ($M = 4.17, SD = 2.63$) in the control condition and 6.20 ($M = 5.95, SD = 2.75$)) in the experimental condition. This corresponds to 42.9% and 66.5% of the score of the optimal strategy, respectively. The presence of the flowchart increased the expected score significantly ($U = 1097.5$, $p < .001$). Moreover, higher click agreement was significantly correlated with higher expected score ($r(117) = .546$, $p < .001$).

These findings indicate that participants did understand the strategy conveyed by the flowchart (higher click agreement), which improved their planning behavior illustrated by the increased expected score.

### 6.3.4 Experiment 4: flowcharts in comparison to performance feedback

In our fourth experiment, knowing that the flowcharts are interpretable for people and that they help them adopt the conveyed strategies, we decided to test how they compare against a real-world alternative. To that end, we measured people's performance in all three environments when they were either assisted by the corresponding flowchart or received performance feedback from an intelligent tutor. The performance feedback condition mimicked how people are taught skills in the real world: the participant receives feedback after he or she has planned and acted. This feedback concerns the correctness of their actions and provides information about the best course of action (see Fig. 10a, b). In the Mouselab MDP, this is equivalent to notifying participants whether they moved correctly and, if not, what move they should have performed.

*Procedure* Participants were randomly assigned to either be assisted by a flowchart or to receive performance feedback in one of the three previously introduced environments (increasing vs. decreasing vs. constant variance), yielding 6 experimental conditions in

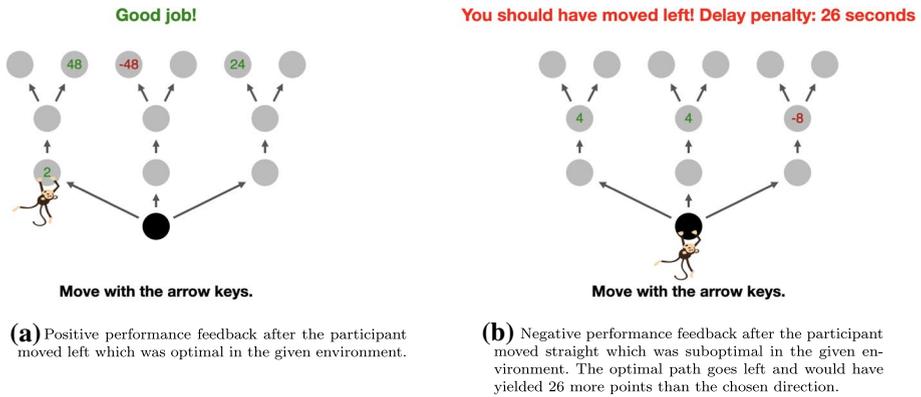

(**a**) Positive performance feedback after the participant moved left which was optimal in the given environment.

(**b**) Negative performance feedback after the participant moved straight which was suboptimal in the given environment. The optimal path goes left and would have yielded 26 more points than the chosen direction.

**Fig. 10** In Experiment 4 real world inspired performance feedback provides feedback on whether the participant's first move was on the best possible action or not total. The flowcharts were not interactive as in Experiment 3, but rather static as in Experiments 1 and 2. The performance feedback was either positive or negative depending on whether the first move of the participant was on the best possible path or not. Positive feedback said "Good job!" in green letters. The negative feedback said "You should have moved<optimal direction>! Delay penalty:<penalty> seconds" in red letters. The game paused during the delay penalty. The duration of the delay was proportional to the difference between the returns of the chosen path and the best path.

Each condition consisted of an introduction to the Mouselab-MDP paradigm including one exploration trial, an introduction to their tutor or the terms used in the flowchart, two practice trials with the tutor, a quiz to test the understanding of the instructions, ten test trials, and a short survey. The flowchart used in the practice block did not convey a reward enhancing strategy, whereas the one in the test block did. To minimize training effects, the 3 training trials were performed in constant variance environments for the decreasing and increasing environment conditions, and in increasing variance environments for the constant variance condition. Exclusively in the constant variance environment conditions, the demanding concept of best paths was conveyed by highlighting nodes on the currently most rewarding path in green, in both the practice trials and the test trials.

We instructed participants in the performance feedback conditions to maximize their score using their own strategy and incentivized them by communicating to pay a performance-dependent bonus. After the experiment, they received 2 cents for each virtual dollar they had earned in the game. We instructed participants in the flowchart conditions to act according to the displayed flowchart and incentivized them by a bonus that was dependent on how often their clicks agreed with the strategy described by the flowchart. After the experiment, they received a bonus proportional to their click agreement score.

*Participants* We recruited 481 people on Amazon Mechanical Turk (average age 37.8 years, range: 18–71 years; 254 female). Each participant received a compensation of $1 plus a performance-dependent bonus up to $0.5. The mean duration of the experiment was 11.4 min. The participants needed 1.8 attempts on average to pass the quiz. We excluded 41 participants (8.5%) who did not perform a single click in the test trials. This resulted in 214 remaining participants for the flowchart conditions and 226 participants for the performance feedback conditions.

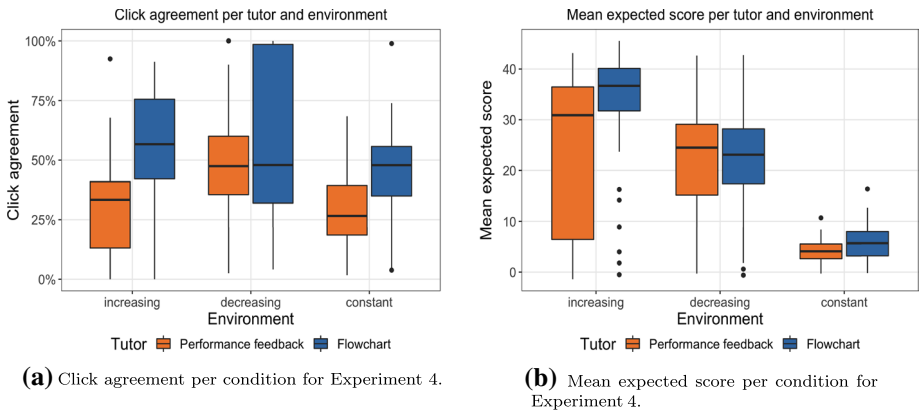

**(a)** Click agreement per condition for Experiment 4.

**(b)** Mean expected score per condition for Experiment 4.

**Fig. 11** Automatically discovered flowcharts improve planning behavior more effectively than real-world inspired performance feedback: **a** Box plot for the click agreement with flowcharts versus performance feedback by environment type (increasing vs. decreasing vs. constant). **b** Box plot for the mean expected score with flowcharts versus performance feedback by environment type (increasing vs. decreasing vs. constant)

*Results* Because the click agreement scores violated the assumptions of normality tested with the Shapiro-Wilk test ($W = .97, p < .001$) and variance homogeneity tested with the Fligner-Killeen test ($\chi^2(5) = 40.01, p < .001$), we decided to perform non-parametric tests: the Mann–Whitney U-test (see the previous experiments) and the robust ANOVA for trimmed means (Wilcox, 2016). Due to the non-normal distribution we also present the median values instead of the means. The median click agreement was 34.9% ($M = 35.5\%, SD = 21.6\%$) in the performance feedback conditions and 49.8% ($M = 52.5\%, SD = 25.6\%$) in the flowchart conditions (see Fig. 11a). The proportion of people who achieved click agreement above 80% increased from 3.98% without the flowchart to 15.88% with the flowchart. Similarly, the proportion of participants who achieved click agreement above 50% increased from 23.89 to 49.53%. We found that there was a significant main effect of whether participants were assisted by a flowchart or received feedback ($\chi^2 = 45.38, p = .001$), a significant main effect of the environment type ($\chi^2 = 25.88, p = .001$), and a significant interaction between the two ($\chi^2 = 6.59, p = .041$). The flowchart was the most effective in improving participant's clicking behavior in the increasing variance environment, while it helped the least in the decreasing variance environment. Concretely, we found a significant effect in the increasing variance environment ($U = 1086, p < .001$) and the constant variance environment ($U = 1475, p < .001$), yet an insignificant effect for the decreasing variance environment ($U = 2328, p = .236$). This can be related to the fact that the strategy of clicking immediate outcomes, conveyed to the participants interacting with the decreasing variance environment, was intuitive. It was also often applied by default in the performance feedback condition. Moreover, the click agreement for the flowchart condition diminished in comparison to Experiment 2.

In the case of the expected scores the assumptions of normality tested with the Shapiro-Wilk test ($W = .91, p < .001$) and variance homogeneity tested with the Fligner-Killeen test ($\chi^2(5) = 112.2, p < .001$) were once again violated. Our comparison thus regards the median values and measures statistical differences with the robust ANOVA and the Mann–Whitney U-test. The median expected score per

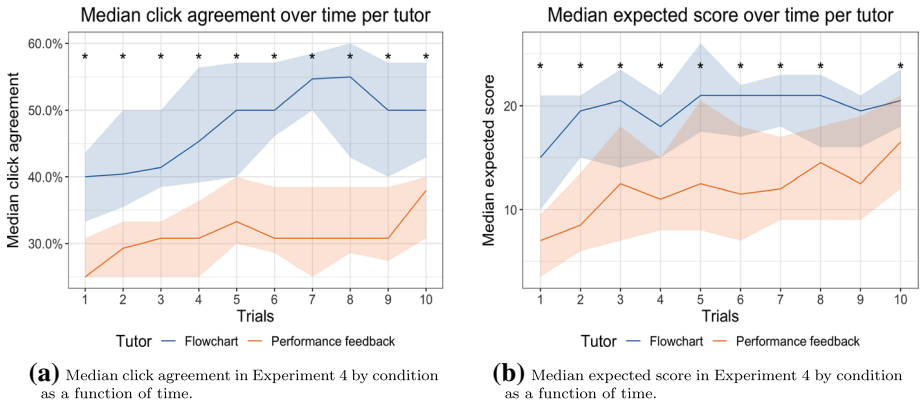

**(a)** Median click agreement in Experiment 4 by condition as a function of time.

**(b)** Median expected score in Experiment 4 by condition as a function of time.

**Fig. 12** Automatically discovered flowcharts improve people's planning behavior more effectively over time than does performance feedback: **a** Learning curve. Median click agreement for the 10 testing trials per tutor (Flowchart vs. Performance feedback). **b** Reward curve. Median expected score for the 10 testing trials per tutor (Flowchart vs. Performance feedback). The shaded areas show the 95% confidence intervals of the medians, whereas the asterisks indicate significant differences for the trial-wise comparisons between the conditions found using the Mann–Whitney U-test ($\alpha = .05$)

trial was 14.25 ($M = 16.84, SD = 13.96$) in the performance feedback conditions and 20.10 ($M = 20.58, SD = 14.21$) in the flowchart conditions. The robust ANOVA for trimmed means revealed that the presence of the flowchart increased the expected score significantly ($\chi^2 = 9.88, p = .003$) (see Fig. 11b). Moreover we found a significant main effect of the environment type ($\chi^2 = 765.99, p = .001$) as well as a significant interaction ($\chi^2 = 8.99, p = .014$). The flowchart was especially effective for participants interacting with the increasing variance environment, whereas it contributed to the scores the least in the decreasing variance condition. Similarly to the click agreement, the effect was significant in the increasing variance environment ($U = 1569, p < .001$) and the constant variance environment ($U = 1957, p = .004$), and insignificant in the decreasing variance environment ($U = 2648.5, p = .937$). Again, this can be attributed to the intuitive and well-performing default strategy for the decreasing variance environment. As revealed by the click agreement scores, this strategy was also applied by the participants in the performance feedback condition.

One could expect that the benefits of flowcharts over performance feedback should decrease over time as people learn from feedback. However, as illustrated in Fig. 12 we found that the benefit of our flowcharts persists with progressing training (see the Supplementary Material for a breakdown of this effect by the type of the environment). In more detail, the average score and the click agreement of the participants in the feedback conditions were constantly lower than the average score and the click agreement of the participants in the flowchart conditions. What is more, Fig. 12a shows that people in both conditions learned the near-optimal strategy increasingly better. This learning effect, which occurred also in the flowchart conditions, paired with a better understanding in the initial phases of the experiment (the flowchart communicates the strategy openly and instantly) prevented the feedback conditions to ever beat the flowchart conditions. Altogether, this shows that performance feedback remains inferior to our flowcharts' guidance irrespective of the accruing time and subjects' experience.

### 6.4 Discussion of findings on improving human decision-making

The results of Experiments 1–3 show that AI-Interpret succeeded to approximate the optimal planning strategies for three different sequential decision problems by simple, human-interpretable decision rules. Presenting these decision rules in the form of flowcharts succeeded to align the way in which people arrived at their decisions more closely with those decision rules. As a consequence, the quality of people's decisions improved. This improvement was statistically significant in the increasing variance environment and in the constant variance environment. The three decision problems differ in how difficult they are for people and in the complexity of strategy that people would have to follow to solve them optimally. Sequential decision problems with decreasing variance are easiest for people because people's intuitive tendency to inspect the immediate rewards first is optimal in this environment. As one would expect based on that, we found that the benefits of our approach were smaller in the decreasing variance environment than in the other two environments where people's intuitive strategies fare poorly. Next, the optimal strategy for the constant variance environment is much more complex than the optimal strategies for the increasing variance environment and the decreasing variance environment. Consequently, we found that people's ability to follow this strategy with or without a flowchart was significantly lower than their ability to follow the optimal strategies for the increasing variance environment and the decreasing variance environment, respectively.

Experiment 4 revealed that the flowcharts our method can generate are significantly better at conveying good decision strategies than the status quo of training people to make better decisions: performance feedback. People who only receive information on the correct course of action through performance feedback planned significantly worse than people who were assisted by flowcharts. In this experiment, we observed worse click agreement for the decreasing and constant variance environments than in the 2 previous studies. For the decreasing variance environment this prevented us from observing a significant difference between the flowchart condition and the feedback condition. This sudden drop was most likely caused by the combination of the following three factors. Firstly, we tested a different sample of Amazon Mechanical Turk users. Secondly, the training block in all of the conditions contained an environment different from the test one, in contrast to the training block in Experiments 1 and 2. Lastly, participants acting in the constant variance environment were tasked with a less demanding training scheme which could have affected their understanding of the flowchart's notions.

Taken together, our findings suggest that interpretable strategy discovery is a promising way to leverage machine learning for designing decision aids. Our approach holds the greatest promise for problems where people's intuitive strategies fare poorly and the optimal strategy is relatively simple. The literature on cognitive biases suggests that there are numerous situations in which people's intuitive strategies perform very poorly (Kahneman et al., 1982) and the literature and heuristic decision-making suggests that there are simple heuristics that people could use to perform much better (Gigerenzer & Gaissmaier, 2011). Additionally, we found that our approach may be more effective than traditional teaching methods, such as performance feedback. This makes interpretable strategy discovery a promising approach for improving human decision making in the real world.

# 7 Benefits of AI-interpret over simpler alternatives

After showing that our relatively complex AI-Interpret algorithm generates human-interpretable decision rules, we now demonstrate that its sophisticated method for selecting a subset of the demonstrations is essential to its success. To achieve this, we compare AI-Interpret against LPP and an ablated version of AI-Interpret. We compare the approaches in terms of the proportion of benchmark problems for which they can find a solution, how consistently they find that solution, and the average performance of found decision rules. Our results support the use of AI-Interpret in the Automatic Discovery of Interpretable Planning Strategies pipeline (see Algorithm 2 and Fig. 1) tested in the previous section. Before delving into this comparison, we briefly introduce the benchmark problems and the baseline methods against which AI-Interpret will be evaluated.

## 7.1 Benchmark problems

To check the performance of our algorithm we tested it on a set of benchmark problems. In each benchmark problem, the algorithm has to find an interpretable description of a reinforcement learning policy. Specifically, we considered the optimal policy for a meta-level MDP (see Definition 2) corresponding to different versions of the planning problem introduced in Sect. 6, found with the dynamic programming method developed by Callaway et al. (2019). In addition to the three planning problems used in Experiments 1–3, the benchmark problems include a fourth type of planning problems where the rewards at the first, second, and third level are drawn from discrete uniform distributions over the values $\{-2, -1, 1, 2\}$, $\{-10, -5, 5, 10\}$, and $\{-20, -10, 10, 20\}$, respectively.

Besides different classes of MDPs, the benchmark problems described in Table 1 also vary the size of the demonstration set from $x = 8$, to $x = 64$, and $x = 128$ trajectories (i.e., sequences of b-c pairs where $b$ is a belief state and $c$ is a computation) starting in the initial belief state ($b_0$) and ending with the termination operation ($\bot$). Each of these trajectories was generated by applying one of the optimal policies to the corresponding meta-level MDP, as described in Sect. 6.2. This resulted in 12 benchmark problems in total (see Table 1). For simplicity, we will refer to them by the number of trajectories and the variance structure of the environment.

## 7.2 Baseline methods

### 7.2.1 LPP

To show the beneficial effects of adding clustering to our algorithm we compared AI-Interpret against the vanilla LPP. In the case of Logical Program Policies, the method has just one shot at the interpretation. The set of demonstrations is split into train and formula-validation sets and based on the supplied DSL, the algorithm searches for a disjunctive normal form formula that provides MAP approximation to the demonstrations. It either finds an interpretable formula, or concludes that the input set of demonstrations is impossible to be described and returns a trivial solution equivalent to the boolean *False*. We use LPP in a loop over maximum conjunction sizes up to input size $d$ to make its results fully comparable to the results of AI-Interpret which also performs this step. The main source of randomness for LPP comes from which demonstrations are assigned to the train set and which are assigned to the formula-validation set. AI-Interpret mitigates these sources of noise by sampling the train and validation sets directly from the clusters it previously creates.

**Table 1** Statistics of the solutions found by the tested interpretation algorithms ran on variable-sized demonstration datasets and Mouselab MDPs with increasing, decreasing, constant, and different variances. The performance metric PERF regards the estimated proportion of the reward a decision tree-induced policy obtains with respect to the optimal policy it is imitating. ENTR specifies the entropy of the distinct results generated over 10 test runs (failure in interpretation is also considered a distinct result). Entropy reflects how stochastic the outputs of the method are across multiple runs on the same dataset. The lower the ENTR measure, the more consistent is the algorithm. SUCC denotes the proportion of succeeded vs. failed runs over the recorded 10 attempts. The error bars enclose 95% confidence intervals. The best results for each benchmark and statistic are bolded

| Benchmark | Method | | |
|---|---|---|---|
| | LPP | Binary-Interpret | AI-Interpret |
| 8 increasing | PERF: 69.35% ± 32.47% ENTR: 1.31 SUCC: 70% | PERF: 69.35% ± 32.47% ENTR: 1.31 SUCC: 70% | PERF: **99.36%** ± **0%** ENTR: **0** SUCC: **100%** |
| 64 increasing | PERF: 58.75% ± 34.31% ENTR: 0.94 SUCC: 70% | PERF: 97.95% ± **0%** ENTR: 0.32 SUCC: **100%** | PERF: **98.01%** ± **0%** ENTR: **0** SUCC: **100%** |
| 128 increasing | PERF: 39.23% ± 34.37% ENTR: 0.67 SUCC: 40% | PERF: 97.99% ± **0%** ENTR: 0.32 SUCC: **100%** | PERF: **99.23%** ± **0%** ENTR: **0** SUCC: **100%** |
| 8 decreasing | PERF: 9.08% ± 21.04% ENTR: **0.32** SUCC: 10% | PERF: 9.08% ± 21.04% ENTR: **0.32** SUCC: 10% | PERF: **93.06%** ± **2.18%** ENTR: 1.37 SUCC: **100%** |
| 64 decreasing | PERF: 0% ENTR: **0** SUCC: 0% | PERF: 0% ENTR: **0** SUCC: 0% | PERF: **94.46%** ± **0%** ENTR: **0** SUCC: **100%** |
| 128 decreasing | PERF: 0% ENTR: **0** SUCC: 0% | PERF: 9.78% ± 21.00% ENTR: 0.32 SUCC: 10% | PERF: **94.66%** ± **0%** ENTR: **0** SUCC: **100%** |
| 8 constant | PERF: 46.94% ± 27.45% ENTR: **1.74** SUCC: 60% | PERF: 46.94% ± 27.45% ENTR: **1.74** SUCC: 60% | PERF: **90.92%** ± **11.40%** ENTR: 2.3 SUCC: **100%** |
| 64 constant | PERF: 0% ENTR: **0** SUCC: 0% | PERF: 8.63% ± 18.52% ENTR: 0.32 SUCC: 10% | PERF: **75.06%** ± **0%** ENTR: **0** SUCC: **100%** |
| 128 constant | PERF: 0% ENTR: **0** SUCC: 0% | PERF: 0% ENTR: **0** SUCC: 0% | PERF: **74.97%** ± **0%** ENTR: **0** SUCC: **100%** |
| 8 different | PERF: 9.39% ± 20.14% ENTR: **0.32** SUCC: 10% | PERF: 9.39% ± 20.14% ENTR: **0.32** SUCC: 10% | PERF: **83.15%** ± **7.35%** ENTR: 2.3 SUCC: **100%** |
| 64 different | PERF: 0% ENTR: **0** SUCC: 0% | PERF: 15.93% ± 23.13% ENTR: 0.64 SUCC: 20% | PERF: **69.8%** ± **0%** ENTR: **0** SUCC: **100%** |
| 128 different | PERF: 0% ENTR: **0** SUCC: 0% | PERF: 7.49% ± 15.91% ENTR: 0.32 SUCC: 10% | PERF: **74.58%** ± **0%** ENTR: **0** SUCC: **100%** |
| Average | PERF: 19.5 ± 6.85% ENTR: **0.2** ± **0.26** SUCC: 26.67% | PERF: 31.1% ± 7.96% ENTR: 0.28 ± 0.14 SUCC: 37.5% | PERF: **87.3%** ± **2.14%** ENTR: 0.49 ± 0.53 SUCC: **100%** |

### 7.2.2 Binary-interpret

To show that the clustering method used in AI-Interpret is the key enabler of our algorithm's success, we compared AI-Interpret against a simpler approach to selecting a subset of the demonstrations, which we call Binary Interpretation (Binary-Interpret). Binary-Interpret uses the principles of the binary search algorithm and accepts the following parameters: aspiration level $\alpha$, tolerance $\delta$, number of rollouts $L$, maximum size $d$, mean expert reward $m$, and an additional parameter *patience*. It also uses one hyperparameter—train-validation split $S$. Binary-Interpret starts by trying to find a formula that satisfies the input constraints using all of the demonstrations. In case of a failure, however, it does not stop but discards half of the demonstrations at random, and tries again on the remaining half of the demonstrations. In case of a success, it increases the size of the demonstration set by the half the size of the previously removed set (if any demonstrations were previously removed) and randomly re-samples the demonstrations. The process continues until the algorithm finds a solution, but fails in the next step or when the difference in size of the currently checked demonstration set and the previous one is equal to or smaller than the *patience* parameter. In each step of Binary-Interpret, the train and formula-validation sets are sampled from each of the demonstrations proportionally to $S = 0.7$. In our tests, we used *patience* $= 8$, meaning that Binary-Interpret stopped when there were only 4 more demonstrations left to consider re-including or removing. The remaining four parameters were shared with AI-Interpret. We used the same values as those listed in Sect. 6.2, namely $\alpha = 0.7$, $\delta = 0.025$, $L = 100{,}000$, $d = 5$ and $m$ dependent on the studied problem, respectively. The pseudo-code detailing Binary-Interpret can be found in the Supplementary Material.

Binary-Interpret is built on the assumption that the more demonstrations there are, the larger is their variety. Since demonstrations can include rare special cases that cannot be captured with the available predicates or cannot be incorporated into the final decision tree, Binary-Interpret checks increasingly smaller sets of demonstrations. It thereby uses the same underlying assumptions as AI-Interpret. We mention it is a clustering method because it performs clustering between demonstrations. Specifically, Binary-Interpret assigns each demonstration to a separate cluster and allows more than one of them to be removed in a single step. In the light of this specification, Binary-Interpret may be viewed as an ablation of AI-Interpret which lacks the component of intelligent clustering.

### 7.3 Quantitative results

The random split of the demonstrations into a training set and a formula-validation set renders the outputs of AI-Interpret, Binary-Interpret and LPP stochastic. Nevertheless, we found the outputs of AI-Interpret to be highly consistent across runs and robust to variations in the set of demonstrations. This made it sufficient to run each algorithm 10 times on our benchmark problems. Despite the error bars of the baselines methods being wide, the amount of data was sufficient to ascertain that the performance of AI-Interpret is significantly better than the performance of the baseline methods. The results showcase that AI-Interpret is very reliable and consistently induces near-optimal decision rules.

To evaluate each method's performance in the context of interpretable strategy discovery, it was inserted into our pipeline for automatically discovering interpretable planning strategies (see Algorithm 2 and Fig. 1). The exact setup and parameters we used for our algorithm, the baselines and Algorithm 2 can be found in Sect. 6.2. In the case of Binary-Interpret and LPP, the pipeline returned just one decision tree since there was no loop over candidate cluster sizes, whereas for our algorithm, the output comprised a set. The exact candidates for the number of clusters that led to the creation of this set were selected by the elbow heuristic applied during the execution of pipeline from Algorithm 2. As detailed in Sect. 6.2, to choose the most interpretable output for each run, we picked the tree with the fewest nodes, and in case of a tie, with the lowest depth. The statistics we use to describe our results in this section are the following:

- Performance ratio (PERF): assuming $f$ is the formula that was turned into a decision tree, this parameter is the average fraction $\frac{m_f}{m_{\pi^*_{\text{meta}}}}$, where $m_f$ and $m_{\pi^*_{\text{meta}}}$ denote the mean reward after 100,000 rollouts of the policy induced by the formula $f$, and the mean reward of the expert policy $\pi^*_{\text{meta}}$, respectively.
- Complexity: the number of nodes of the output tree.
- Support: the mean proportion of state-action pairs which were used to find the most interpretable result.
- Entropy (ENTR): the entropy of solutions (including a failure solution) generated across 10 runs in total. Lower values indicate that the method is more reliable because its outputs are less variable.
- Success rate (SUCC): the proportion of times the algorithm generated a formula out of 10 runs in total. Measuring the success rate helps us to numerically capture the effectiveness of each method.

When LPP is unable to find a decision rule it returns *false* which entails no planning. Thus, when any of the three methods is unable to find any formula that is consistent with (any subset of) the demonstrations, the resulting decision strategy uses zero information about the environment. The performance ratio of this no-planning strategy is 0 because the expected value of the reward distributions we used in our benchmark problems is always 0. Runs that did not output any decision tree counted as unsuccessful and were considered in the calculation of the entropy metric. Table 1 presents the performance of all methods on each of the 12 benchmark problems in terms of three metrics: performance ratios with 95% confidence intervals, entropy, and success ratios. Rows correspond to different benchmarks and columns relate to the algorithms. Additional statistics are reported in the main text.

### 7.3.1 Benchmark evaluation

By inspecting the evaluation presented in Table 1, we see that AI-Interpret consistently managed to discover simple, high-performing decision rules.

While AI-Interpret succeeded to find an interpretable decision rule on every single one of its 120 runs on the benchmark problems, LPP and Binary-Interpret succeeded on only 26/120 and 40/120 runs on the benchmark problems, respectively. Two $\chi^2$-tests for contingency tables confirmed that our algorithm succeeds in finding interpretable descriptions significantly more often than Binary-Interpret ($\chi^2(4) = 31.82$, $p < .0001$) and LPP ($\chi^2(4) = 31.82$, $p < .0001$). Concretely, LPP output a non-trivial solution in 18/30 cases for the increasing variance benchmarks, 1/30 cases for the decreasing variance benchmarks, 6/30 cases for the constant variance benchmarks, and 1/30 cases for the different variance benchmarks. Binary-Interpret output a formula 27/30 times for the increasing variance benchmarks, 2/30 times for the decreasing variance benchmarks, 7/30 times for the constant variance benchmarks and 4/30 times for the different variance benchmarks.

Secondly, we compared AI-Interpret with the baselines on the basis of the performance ratio (PERF). On average across all benchmark problems, the decision rules induced by AI-Interpret achieved 87.3% ± 2.14% of the return of the optimal metalevel policy. By contrast, the performance ratios of LPP and Binary Interpret were merely 19.5% ± 6.85% and 31.1% ± 7.96%, respectively. Mann–Whitney U-tests[1] confirmed that these differences are statistically significant (AI-Interpret vs. Binary-Interpret: $U = 3173.0$, $p < .0001$; AI-Interpret vs. LPP: $t = 2061.0$, $p < .0001$). As shown in Table 1, the performance benefit of AI-Interpret was consistently present across all of the benchmark problems.

The entropy metrics shown in Table 1 suggest that AI-Interpret always outputs the same solution when 64 or 128 demonstrations are provided but is less stable when only 8 demonstrations are provided. On average, the descriptions generated by Binary-Interpret and LPP had a reasonably low entropy too. But while AI-Interpret achieved consistency by consistently finding good decision rules (100% success rate, 85.16% lower confidence bound on performance), LPP and Binary-Interpret consistently failed to find any solution on the majority of the benchmark problems (Binary-Interpret with 31.3% ± 29.7% success rate and 39.05% upper confidence bound on performance; LPP with 13.75% ± 18.5% success rate and 26.35% upper confidence bound on performance).

Knowing that our algorithm is largely superior than both LPP and Binary-Interpret, we moved to descriptively describing the formulas it finds (or decision trees if used in Algorithm 2). In this way, we noted that the smallest decision trees induced from AI-Interpret's output had merely 1 node, whereas the biggest ones needed 8 nodes. Still, the mean complexity was very low and equaled 2.75 ± 0.31. All but one of the found interpretable descriptions were discovered on a modified input dataset that excluded some of the demonstrations; on average AI-Interpret had a support of 59.42% ± 3.44% of all state-action pairs. Moreover, variations within the environments were not significantly large. For constant variance benchmark problems the support equaled 52.43% ± 6.08%; for different variance problems it was 48.64% ± 5.47%; for the decreasing variance problems it was 52.83% ± 1.56%; and for the increasing variance problems it was 83.77% ± 4.22%. These measures indicate that AI-Interpret chose the demonstration set proportion adaptively depending on the environment. Furthermore, inspecting the clustering value as a function of the number of clusters revealed that the elbow heuristic is a useful criterion for choosing that number to work well with AI-Interpret (see Supplementary Material).

These findings highlight that AI-Interpret is clearly superior to LPP and Binary-Interpret. Intelligent clustering enables it to find solutions when LPP and Binary-Interpret fail. The performance ratios of these solutions indicate that AI-Interpret discovers policies of high quality. The low entropy in terms of the different outputs and high success rate pinpoint that clusters indeed capture similar demonstrations, and make the output reliable and trustworthy. We therefore conclude that the innovations of AI-Interpret were critical to the success of interpretable strategy discovery at improving human decision-making reported in Sect. 6 and that our algorithm is ready to be applied to other reinforcement learning problems requiring human-interpretability.

---

[1] We chose this non-parametric test because the performance ratios of Binary-Interpret and LPP violated the normality assumptions of common parametric statistical methods.

## 8 General discussion and future directions

Decision aids, such as decision trees and flowcharts, help professionals (e.g., medical doctors) make better decisions by guiding them through a more systematic decision strategy that prioritizes the most important information. Recent advances in cognitive science make it possible to leverage reinforcement learning methods to discover optimal versions of such strategies automatically (Callaway et al., 2019; Griffiths et al., 2019; Lieder and Griffiths, 2020; Lieder et al., 2017).

In this article, we extended a reinforcement learning method that automatically discovers near-optimal decision-making strategies through an addition of interpretable RL algorithm AI-Interpret. This extension enabled the mentioned method to automatically generate near-optimal decision aids instead of outputting compound RL policies, as it has done before. The pipeline for automated discovery of interpretable strategies takes four main steps: (1) it models a decision problem as a metalevel MDP, (2) it solves for the optimal metalevel policy, (3) it interprets this policy with AI-Interpret, and (4) it turns the resulting formula to a human-interpretable flowchart. Our proof-of-concept demonstrations showed that the decision-aids generated by this method can improve people's planning strategies and the quality of the resulting decisions more effectively than conventional performance feedback. While AI-Interpret builds on a promising Bayesian program induction approach to imitation learning (i.e., LPP, Silver et al., 2019), we found that its innovations are in fact critical. The original version of LPP and simpler extensions were not robust enough to tackle the real-world challenges of *approximating* complex, stochastic, and irregular policies with simple decision rules that can be readily understood and applied by people. AI-Interpret achieves this robustness by clustering the set of demonstrations and identifying the largest possible set of behaviors that can be captured by an easily comprehensible logical formula. The results of our quantitative experiments clearly indicate a beneficial effect of clustering the set of demonstrations. The ablated version of AI-Interpret managed to find reliable decision rules only for one out of four types of sequential decision problems, whereas AI-Interpret consistently found well-performing, simple rules for all of them.

These findings indicate that AI-Interpret is an important step towards leveraging reinforcement learning to boost people's decision-making skills in real life. This illustrates how interpretable machine learning can be used to help people perform better instead of replacing them entirely.

### 8.1 Directions for future work

AI-Interpret is a very general method with a broad range of possible real-life applications. These applications include improving human decision-making, understanding the decisions of artificial intelligent systems, and understanding human decisions. Each of these applications can be pursued in a wide range of domains including planning, decision-making, reasoning, vision, robotics, and learning.

Future work will extend our approach to helping people make better decisions to increasingly more realistic scenarios, such as purchasing, hiring, (college) admissions, investing, and medical diagnosis. For example, one of the directions we plan to explore in the near future is discovering human-interpretable decision strategies for multi-alternative risky choice problems that model real-life investment decisions (e.g., Rieskamp & Otto, 2006; see Fig. 13).

**Fig. 13** Investment task used by Rieskamp and Otto (2006). To decide which company to invest in, the decision-maker can compare the companies on multiple different attributes, such as market share, financial flexibility, image, efficiency, management, etc. AI-Interpret can be applied to discover optimal decision strategies for this investment task as part of our automatic planning strategy discovery pipeline (see Fig. 1)

A natural extension of multi-alternative risky choice is the topic of product selection illustrated in Fig. 2a). Furthermore, we will also apply AI-Interpret to partially automate the process of scientific discovery in cognitive science by assisting cognitive scientists in their efforts to derive people's decision strategies from the order in which they inspect different pieces of information (see Fig. 2a). Even though we pointed out that Mouselab-MDP used as a benchmark for our tests poses a challenge to the standard methods due to its enormous belief state-space, other works in the field (e.g. Verma et al., 2018; Verma et al., 2019) deal with complex continuous environments. Solving such problems may require deep learning. Future work should thus also explore applying our approach to explain the decisions of deep neural networks that perform at a super-human level (e.g., Mnih et al., 2015; Silver et al., 2018; Silver et al., 2017) and to transfer their expertise to people.

To establish a solid foundation for these real-world applications, future research will rigorously analyze the AI-Interpret algorithm in the theoretical framework of statistical learning theory (Vapnik, 2013). We also plan to explore translating the decision tree that our method generates into a program in linear temporal logic (Camacho & McIlraith, 2019; Vazquez-Chanlatte et al., 2018) that specifies which operation should be performed next. We predict that such a programmatic representation will be much more helpful for people than flowcharts for determining whether a given planning operation is consistent with the recommended strategy. Finally, we are also actively working to address the issue of scalability of the whole pipeline for Interpretable Strategy Discovery 2. Standard methods for solving metalevel MDPs that model real-life problems in our pipeline cannot handle environments larger than 20–30 nodes. New approaches we are developing can find optimal policies for increasingly realistic planning problems that require thousands of nodes (Consul et al., 2021).

Interpretable flowcharts can be used not only as decision aids but also for teaching effective decision strategies. Existing cognitive tutors teach decision strategies primarily by giving people feedback about how they make their decisions while they practice decision-making in a simulated environment (Lieder et al., 2019; Lieder et al., 2020). Since this pedagogical approach primarily relies on implicit learning, people's conscious understanding of the taught strategy tends to be limited to its application in the training environment (Lieder et al., 2020). Interpretable flowcharts, by contrast, represent strategies in general terms that are directly applicable to decision-making in real-life. Adding interpretable flowcharts to cognitive tutors might therefore make it much easier for people to transfer what they were taught to decision-making in everyday life. This makes augmenting cognitive tutors with AI-Interpret and Algorithm 2 another promising direction for future work.

In the long run, this line of research may lead to deep insights into decision-making, clever cognitive strategies, and practical tools that help people make better decisions in many important real-life situations. In this way, advances in artificial intelligence can enhance human intelligence instead of replacing people. This is an important antidote to people losing their jobs to robots and algorithms.

**Supplementary Information** The online version contains supplementary material available at https://doi.org/10.1007/s10994-021-05963-2.

**Acknowledgements** This work was supported by the German Federal Ministry of Education and Research (BMBF): Tübingen AI Center, FKZ: 01IS18039B.

**Funding** Open Access funding enabled and organized by Projekt DEAL.

**Publisher's Note** Springer Nature remains neutral with regard to jurisdictional claims in published maps and institutional affiliations.

## Authors and Affiliations

**Julian Skirzyński[1]**  **· Frederic Becker[1] · Falk Lieder[1]**

Frederic Becker
frederic.becker@tuebingen.mpg.de

Falk Lieder
falk.lieder@tuebingen.mpg.de

[1] Max Planck Institute for Intelligent Systems, Tübingen, Germany